\documentclass{article}

\usepackage{microtype}
\usepackage{graphicx}
\usepackage{subcaption}
\usepackage{booktabs} 
\usepackage{microtype}
\usepackage{hyperref}
\usepackage{url}
\usepackage{xurl}
\usepackage{booktabs}
\usepackage{graphicx}
\usepackage{float}
\usepackage{xcolor}
\usepackage{amsmath}
\usepackage{amssymb}
\usepackage{algorithm}
\usepackage{algpseudocode}
\usepackage{subcaption}
\usepackage{wrapfig}
\usepackage{fvextra}
\usepackage{comment}
\usepackage{siunitx}
\usepackage{multirow}
\usepackage{listings}
\usepackage[T1]{fontenc}

\usepackage{hyperref}

\newcommand{\tsc}[1]{\textsc{#1}}
\newcommand{\mc}[1]{\mathcal{#1}}

\usepackage[accepted]{icml2026}

\usepackage{amsmath}
\usepackage{amssymb}
\usepackage{mathtools}
\usepackage{amsthm}

\usepackage[capitalize,noabbrev]{cleveref}

\theoremstyle{plain}

\theoremstyle{definition}

\theoremstyle{remark}

\usepackage[textsize=tiny]{todonotes}

\icmltitlerunning{MKEvolve: A Modular Multi-Agent Framework for Kernel Code Generation}

\begin{document}

\twocolumn[
  \icmltitle{MKEvolve: A Modular Multi-Agent Framework for Kernel Code Generation}



  \icmlsetsymbol{equal}{*}
  \icmlsetsymbol{workdone}{\textdagger}

  \begin{icmlauthorlist}
    \icmlauthor{Jason Yoo}{ubc,workdone}
    \icmlauthor{Rajarshi Saha}{aws}
    \icmlauthor{Shaowei Zhu}{aws}
    \icmlauthor{Tao Yu}{aws}
    \icmlauthor{Wei Tang}{aws}
    \icmlauthor{Youngsuk Park}{aws}
  \end{icmlauthorlist}

  \icmlaffiliation{ubc}{University of British Columbia}
  \icmlaffiliation{aws}{Amazon Web Services}

  \icmlcorrespondingauthor{Jason Yoo}{jasonyoo0116@gmail.com}
  \icmlcorrespondingauthor{Rajarshi Saha}{sahrajar@amazon.com}
  \icmlcorrespondingauthor{Youngsuk Park}{pyoungsu@amazon.com}

  \icmlkeywords{Machine Learning, ICML}
  \vskip 0.3in
]



\printAffiliationsAndNotice{\textsuperscript{\textdagger}Work done during an internship at Amazon Web Services.}

\begin{abstract}

Despite rapid progress in LLM-based code generation, writing correct and performant kernels for hardware accelerators remains a key bottleneck in scaling modern ML workloads.
We present \tsc{MKEvolve} ({\bf M}odular {\bf K}ernel {\bf E}volve), a framework that iteratively co-evolves a modular decomposition of complex PyTorch modules and the LLM-generated kernel for each submodule, refining the decomposition by splitting and fusing across iterations while independently improving each subkernel via LLM-driven beam search.
The resulting kernels are programmatic compositions of independently verified subkernels, making them configurable (subkernel implementations are swappable), interpretable (errors and speedups are traceable to specific subkernels), and readily adaptable to related model architectures.
Experiments with Triton on KernelBench L2 and L3, spanning multi-operator sequences and full model architectures, show that \tsc{MKEvolve} improves both correctness and speedup over end-to-end direct synthesis baselines while reducing LLM token usage by up to 35\%.

\end{abstract}

\section{Introduction}

The rapid growth of modern machine learning workloads has sharply increased demand for efficient kernels that improve hardware utilization and throughput, directly reducing inference latency and energy cost. 
However, writing such kernels remains difficult even for domain experts since high performance depends on carefully orchestrating parallelism, memory movement, tiling, fusion, and numerical behavior in ways that are tightly coupled to both the target hardware and the specific structure of the workload. 
As a result, kernel development is often time-consuming, brittle, and inaccessible to most practitioners.


Recently, LLM-driven kernel generation has emerged as a promising way to reduce this burden, with \citet{kernelllm2025, li2025autotriton, woo2025tritonrl, zhang2025accelopt, wang2025kernelfalcon} automatically synthesizing kernels from framework-level specifications such as \citet{ouyang2025kernelbench}.
However, the majority of existing approaches treat the kernel as a single monolithic artifact throughout the improvement loop, creating two key limitations: {\rm (i)} repeatedly optimizing the entire kernel can be inefficient in terms of improvement effort and token usage since many useful improvements are local and affect only a small portion of the computation
and {\rm (ii)} the resulting solutions are often complex and difficult for both non-experts and LLMs to interpret, debug, or adapt. 

While these challenges apply broadly to LLM-based kernel generation, this work focuses on Triton, a distinct and timely setting.
Unlike CUDA, which benefits from abundant high-quality training data enabling effective post-training approaches such as CudaAgent~\citep{dai2026cuda}, Triton lacks sufficient high-quality data.
As a result, post-training methods are less effective and often exhibit reward hacking, as observed in TritonRL~\citep{woo2025tritonrl}, making agentic test-time methods the current frontier for Triton synthesis.


This work introduces \tsc{MKEvolve}, a modular kernel generation framework that operates on kernels structured as compositions of subkernels, each solving an explicitly defined subproblem.
Rather than optimizing a kernel end-to-end, \tsc{MKEvolve} repeatedly decomposes a complex PyTorch module into smaller submodules, independently improves kernels for each via LLM-driven beam search, and programmatically composes them into a complete implementation. 
This modularization makes the search space more structured and tractable: local improvements can target where they matter most, alternative implementations can be explored at the submodule level, and failures can be traced to specific subkernels rather than an opaque monolithic program. 
The resulting kernels are not only more reliable but also more interpretable and easier for engineers to modify, reuse, and extend to related architectural variants with common submodules.

\begin{figure*}[t]
    \centering
    \includegraphics[width=0.9\linewidth]{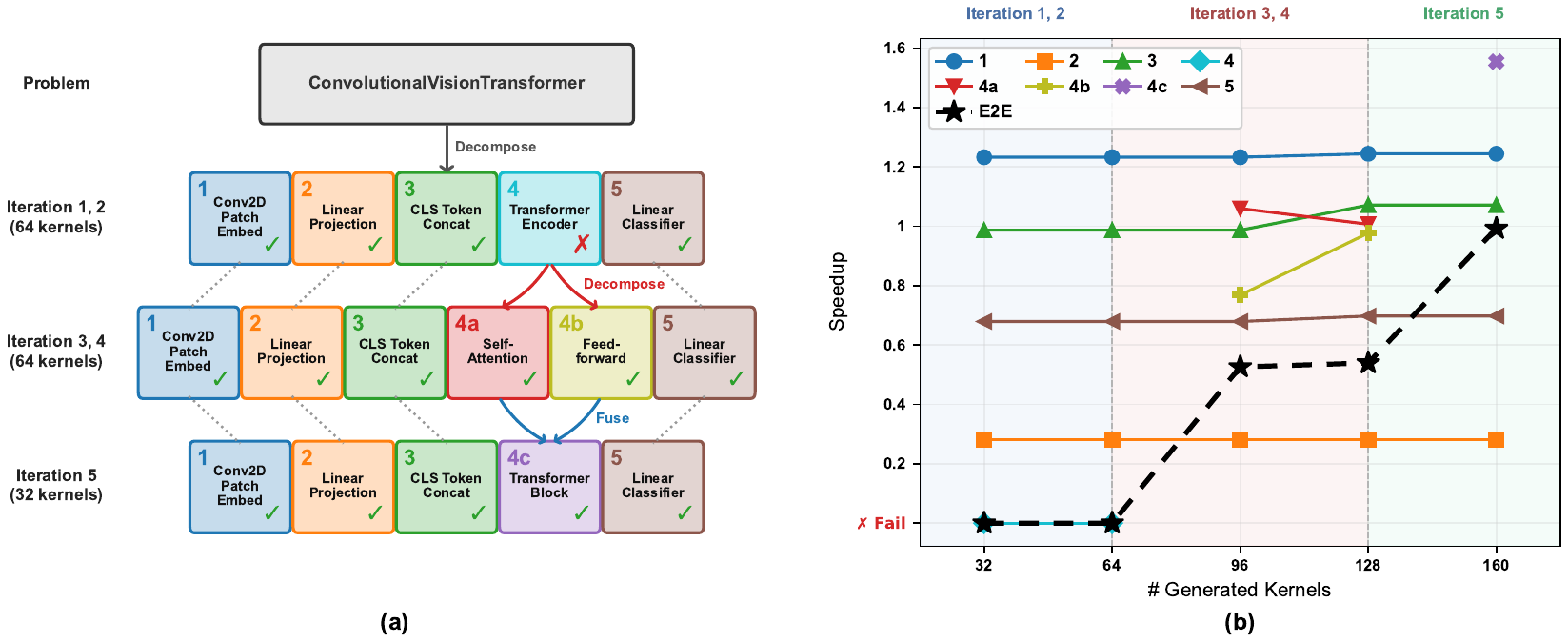}
    \caption{
    \tsc{MKEvolve} kernel synthesis on the KernelBench L3 ConvolutionalVisionTransformer task for 5 outer loop iterations.
    \textbf{Left}: Subproblem structure evolution during refinement, with checkmarks and crosses respectively denoting correct and invalid subkernels.
    \tsc{MKEvolve} iteratively decomposes the subproblem structure until all subproblems admit correct subkernels, then considers subproblem fusion.
    \textbf{Right}: Subkernel (Lines 1-5) and end-to-end (\textbf{E2E}) kernel speedup relative to \textit{torch.compile} during refinement.
    End-to-end kernel speedup increases from failure to 0.53 after the Transformer Encoder subproblem (4) is decomposed into Self-Attention (4a) and Feed-forward (4b), then increases significantly again when (4a) and (4b) are merged into the Transformer Block subproblem (4c).
    }
    \label{fig:mainfigure}
    \vspace*{-0.5em}
\end{figure*}

Our contributions are threefold: 
{\rm (i)} we introduce \tsc{MKEvolve}, the first kernel code generation framework that iteratively co-evolves a modular decomposition and independently optimized subkernels, targeting correctness while achieving speedup; 
{\rm (ii)} we show that this compositional structure yields kernels that are interpretable, configurable, and readily transferable to architecturally related workloads; and 
{\rm (iii)} we evaluate \tsc{MKEvolve} with Triton on KernelBench L2 and L3, yielding improvements in both correctness and runtime over direct synthesis baselines while reducing LLM token usage by up to 35\%.
Taken together, our results establish modularity as a valuable organizing principle for LLM-driven kernel generation.\footnote{Code will be released at \url{https://github.com/amazon-science/ModularKernelEvolution}.}

\section{Related Work}

\paragraph{Post-training} Recent works improve kernel generation by fine-tuning LLMs on kernel code datasets. 
KernelLLM~\citep{kernelllm2025} and Concur~\citep{kong2025concur} use supervised tuning on curated datasets, while Kevin~\citep{baronio2025kevinmultiturnrlgenerating} applies reinforcement learning (RL) for correctness and performance. 
CUDA Agent \citep{dai2026cuda} further advances this idea by training LLMs via large-scale agentic RL, combining data synthesis and execution feedback in a skill-augmented environment.
AutoTriton~\citep{li2025autotriton} and TritonRL~\citep{woo2025tritonrl} combine both; TritonRL also adds cheating detection, which we adopt in our pipeline.
However, these require direct access to model weights, making them unusable with closed models that only expose API access.
Moreover, \tsc{MKEvolve} is complementary, operating purely at inference time and is able to leverage stronger post-trained LLMs.



\paragraph{Inference scaling} A parallel line of work focuses on improving kernel quality at inference time by scaling LLM computation. 
At the simplest end, MultiKernelBench~\citep{wen2025multikernelbench} studies one-shot prompting across platforms, while \citet{lange2025towards} explore parallel sampling of independent candidates. 
More structured search strategies include beam search in AccelOpt~\citep{zhang2025accelopt} and AutoComp~\citep{hong2025autocomp}, and evolutionary methods in AlphaEvolve~\citep{novikov2025alphaevolve} and Avo~\citep{chen2026avo}, which iteratively refine candidate kernels. 
Multi-agent systems such as Astra~\citep{wei2025astra}, KernelFalcon~\citep{wang2025kernelfalcon}, and KernelEvolve~\citep{liao2025kernelevolve} partitioning the generation process across specialized agents. 
\tsc{MKEvolve} falls in this category but is distinguished by its iterative co-evolution of the decomposition structure and subkernel implementations, with beam search applied independently at the submodule level.

A closely related work is KernelFalcon \citep{wang2025kernelfalcon}, a multi-agent framework that constructs an LLM-generated JSON representation approximating the PyTorch compute graph, synthesizes subkernels for each subgraph using LLM-generated correctness tests, and composes them into an end-to-end kernel via an LLM.
In contrast, \tsc{MKEvolve} introduces a key design distinction: it programmatically composes subkernels directly within the end-to-end kernel throughout the synthesis process in a verifiable manner.
This enables both reliable reuse and flexible substitution of optimized subkernels across tasks.
%
%

\begin{figure*}[t]
    \centering
    \includegraphics[width=0.825\linewidth]{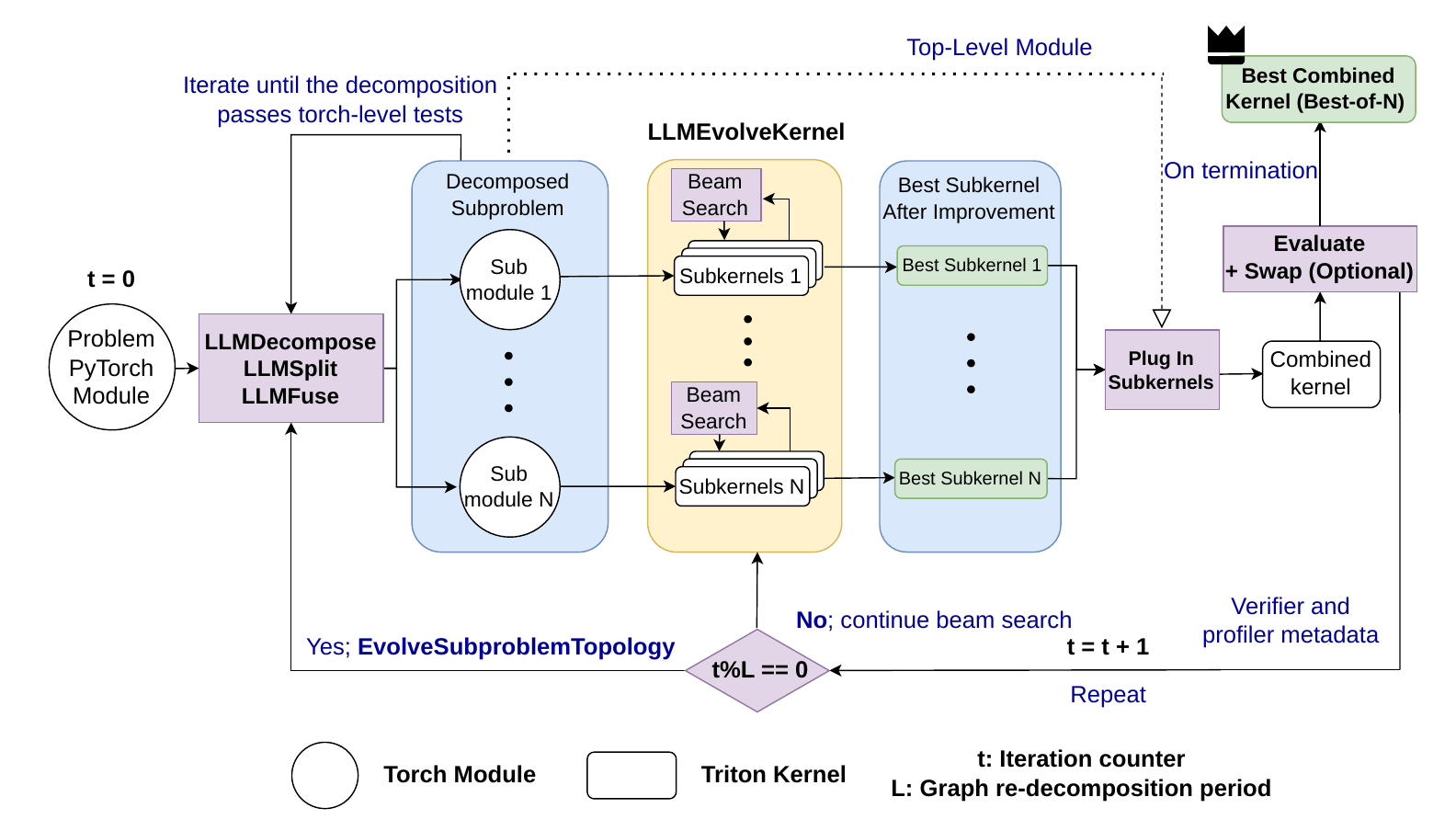}
    \caption{
    MKEvolve \Cref{alg:mkevolve} Visualization.
    }
    \label{fig:method_diagram}
    \vspace*{-0.5em}
\end{figure*}

Furthermore, \tsc{MKEvolve} adopts an iterative optimization strategy, evolving subkernels using beam search rather than stopping at the first correct solution and refining the subproblem decomposition itself, resulting in more efficient kernel implementations.
Finally, \tsc{MKEvolve} relies on a programmatic evaluation pipeline with cheating detection and strict correctness criteria, instead of LLM-generated tests, improving robustness and reliability.

\section{Methodology}

\begin{algorithm*}[t]
\caption{MKEvolve}
\label{alg:mkevolve}
\begin{algorithmic}[1]
\Require PyTorch module $M$, Verifier $\mathcal{V}$, Iterations $T$, Refinement period $L$, Threshold $\tau$
\Ensure Best subkernels $\mathcal{K}^*$, Best top-level module $S^*$, Best speedup $r^*$

\State $\mathcal{P}, S \gets \text{LLMDecompose}(M)$ \Comment{Initial subproblems and top-level module}
\State $\mathcal{K} \gets [\mathcal{K}_i]_{i=1}^{|\mathcal{P}|}$, \quad\quad $\mathcal{K}_i = (\emptyset, 0)$ \Comment{Initial subkernels and scores list}

\For{$t = 1 \dots T$}
    \State $\mathcal{P}, \mathcal{K}, S \gets \text{EvolveSubproblemTopology}(\mathcal{P}, \mathcal{K}, t, L)$ \Comment{See Algorithm 2}
    \State $[b_1, \dots, b_{|\mathcal{P}|}] \gets \text{AllocateLLMBudget}(\mathcal{K})$ \Comment{Prioritize based on runtime/error}
    
    \For{$i = 1 \dots |\mathcal{P}|$}
        \State $\mathcal{K}_i \gets \text{LLMEvolveKernel}(\mathcal{P}_i, \mathcal{K}_i, b_i, \mathcal{V})$ \Comment{Local subkernel improvement}
    \EndFor
    
    \State $r \gets \text{Evaluate}(S, \mathcal{K}, \mathcal{V})$ \Comment{End-to-end kernel verification and profiling}
    \If{$r \geq r^*$}
        \State $\mathcal{K}^*, S^*, r^* \gets \mathcal{K}, S, r$
    \EndIf
\EndFor

\If{swapping is enabled}
    \State $\mathcal{K}^* \gets \text{Swap}(\mathcal{K}^*, \tau)$ \Comment{Substitute poor subkernels with alternatives}
\EndIf
\State \Return $\mathcal{K}^*, S^*$
\end{algorithmic}
\end{algorithm*}


\begin{algorithm}[t]
\caption{EvolveSubproblemTopology ($\mathcal{P}, \mathcal{K}, t, L$)}
\label{alg:refine}
\begin{algorithmic}[1]
\If{$t > 0$ and $t \bmod L = 0$} \Comment{Periodic refinement}
    \If{there are failing subkernels $\mathcal{K}_{\text{fail}} \subset \mathcal{K}$}
        \State $\mathcal{P}, S \gets \text{LLMSplit}(\mathcal{P}, \mathcal{K}_{\text{fail}})$
    \ElsIf{fusion is enabled}
        \State $\mathcal{P}, S \gets \text{LLMFuse}(\mathcal{P}, \mathcal{K})$
    \EndIf
\EndIf

\If{local and global correctness checks do not agree}
    \State $\mathcal{P}, S \gets \text{TotalFuse}(\mathcal{P})$ \Comment{Monolithic problem}
\EndIf

\State $\mathcal{K} \gets \text{FilterKernels}(\mathcal{P}, \mathcal{K})$ \Comment{For deleted subproblems}
\State $\mathcal{K} \gets \text{InitEmptyKernels}(\mathcal{P}, \mathcal{K})$ \Comment{For new subproblems}

\State \Return $\mathcal{P}, \mathcal{K}, S$
\end{algorithmic}
\end{algorithm}

\Cref{alg:mkevolve} outlines the core \tsc{MKEvolve} workflow.
\tsc{MKEvolve} produces a Python \textit{codebase} that implements the given PyTorch module in a modular, kernelized form.
This codebase consists of a set of files $\mc{K}$, each implementing a kernel for a subproblem derived from the original module $M$, along with a single top-level module file $S$ that serves as the entry point and orchestrates the subkernels in $\mc{K}$.
A representative \tsc{MKEvolve} codebase is provided in Appendix \ref{appendix:cvtcode}.
We now describe each algorithm component in detail.


\paragraph{LLMDecompose agent} decomposes the initial problem into subproblems $\mc{P}$ and a top-level module $S$.
The agent is instructed to decompose the original problem into subproblems when fully fused kernels are difficult to implement correctly, while maximizing intra-subproblem fusion opportunities and delegating all mathematical computation to the subproblems.
Its input is a Python file containing the problem PyTorch module along with input generation functions for module's \texttt{\_\_init\_\_} and \texttt{forward} methods, and its output is a Python codebase comprised of subproblem PyTorch module files and a top-level module file that chain them into a functionally identical implementation of the original PyTorch module.
During execution, the agent launches workers that attempt decomposition in parallel via sequential scaling, and returns as soon as any worker produces a solution that passes programmatic checks using PyTorch hooks, including but not limited to:
\begin{itemize}
    \item Functional correctness: Verifying that the original module and the decomposed codebase produce identical outputs on the same input.
    \item \texttt{forward} shape coverage: Ensuring each subproblem's \texttt{forward} input generator covers all input shapes used by the top-level module.
    \item \texttt{\_\_init\_\_} input coverage: Ensuring each subproblem's \texttt{\_\_init\_\_} input generator covers all \texttt{\_\_init\_\_} initialization inputs used by the top-level module.
\end{itemize}
These checks ensure that the generated subproblems fully and exclusively cover the initialization and forward inputs required by the generated top-level module. 
%
Otherwise, a fallback top-level module is returned that wraps the original module as the sole submodule.

\paragraph{LLMEvolveKernel agent} generates, debugs, and optimizes Triton kernels for the given (sub)problem for $b$ inner loop iterations.
We use beam search, where the reward is kernel speedup vs \textit{torch.compile}, and incorrect kernels have a speedup of 0.
LLMEvolveKernel agent's verifier comprises of TritonRL-based cheating detector \citep{woo2025tritonrl} and the KernelBench repository's CUDA-stream evaluation pipeline.

\paragraph{LLMSplit agent} attempts to split a list of provided subproblems into exactly 2 PyTorch subproblems each, and restructure the top-level module to account for this change.
If this fails, the previous subproblem and top-level module are returned.

\paragraph{LLMFuse agent} attempts to fuse any number of subproblems PyTorch modules into larger modules at the LLM's discretion, and restructure the top-level module to account for this change.
If this fails, the previous subproblem and top-level module are returned.
We denote \tsc{MKEvolve} with LLMFuse enabled as \tsc{MKEvolve (Fuse)}.

\paragraph{AllocateLLMBudget} allocates the iteration's LLM call budget assigned to LLMEvolveKernel agents tasked with solving different subproblem kernels.
Specifically, if there are any subproblem(s) that are new or has incorrect Triton kernel implementation, it evenly distributes the LLM call budget between them.
Otherwise, it allocates the LLM call budget to each subproblem agent proportionally to the subproblem kernel's runtime observed during the last end-to-end evaluation. 

\paragraph{Evaluate} verifies and profiles the end-to-end kernel implemented by the top-level module against the original problem code, with the verifier used by LLMEvolveKernel.
The output reward is speedup against \textit{torch.compile}, with incorrect kernels receiving 0 reward.

\paragraph{Swap} iterates through all subkernels and programmatically replaces subkernels whose speedup relative to \textit{torch.compile} is below the swap threshold $\tau$ with alternative implementations (ex. PyTorch, subkernels from a more powerful LLM).
Swap is not triggered if there are fewer than two subkernels or if all subkernels are slower than \textit{torch.compile}.
We denote \tsc{MKEvolve} with Swap enabled as \tsc{MKEvolve (Swap)}.

\begin{table*}[t]
\centering
\small

\begin{subtable}{0.775\linewidth}
\centering
\caption{Claude 4.5 Opus metrics on 100 KernelBench level 2 problems.}
\resizebox{0.92\linewidth}{!}{%
\begin{tabular}{lccccc}
\toprule
\textsc{Method} & Correct (↑) & Fast$_{0.5}$ (↑) & Fast$_{1}$ (↑) & Fast$_{2}$ (↑) & \# Tokens \\
\midrule
Parallel Scaling & 0.68 & 0.46 & 0.03 & 0.01 & $1.3 \times 10^6$ \\
Beam Search & 0.96 & 0.72 & 0.36 & 0.07 & $2.0 \times 10^6$ \\
MKEvolve & \textbf{0.99} & \textbf{0.77} & 0.49 & \textbf{0.09} & $1.7 \times 10^6$ \\
MKEvolve (Fuse) & 0.98 & 0.76 & \textbf{0.54} & \textbf{0.09} & $1.9 \times 10^6$ \\
\midrule
KernelFalcon & 0.93 & 0.22 & 0.00 & 0.00 & $0.2 \times 10^6$ \\
MKEvolve (PostFuse) & 0.99 & 0.80 & 0.55 & \textbf{\underline{0.10}} & $1.8 \times 10^6$ \\
MKEvolve (Swap) & \textbf{\underline{1.00}} & \textbf{\underline{0.87}} & \textbf{\underline{0.58}} & 0.09 & $1.7 \times 10^6$ \\
\bottomrule
\end{tabular}%
}
\label{tab:c4o_l2}
\end{subtable}

\vspace{0.75em}

\begin{subtable}{0.775\linewidth}
\centering
\caption{Claude 4.5 Opus metrics on 50 KernelBench level 3 problems.}
\resizebox{0.92\linewidth}{!}{%
\begin{tabular}{lccccc}
\toprule
\textsc{Method} & Correct (↑) & Fast$_{0.5}$ (↑) & Fast$_{1}$ (↑) & Fast$_{2}$ (↑) & \# Tokens \\
\midrule
Parallel Scaling & 0.72 & 0.32 & 0.14 & 0.06 & $1.8 \times 10^6$ \\
Beam Search & 0.88 & 0.64 & 0.26 & \textbf{\underline{0.10}} & $3.0 \times 10^6$ \\
MKEvolve & \textbf{\underline{0.94}} & \textbf{0.74} & \textbf{0.34} & 0.08 & $2.1 \times 10^6$ \\
MKEvolve (Fuse) & 0.92 & 0.68 & 0.32 & 0.08 & $2.9 \times 10^6$ \\
\midrule
KernelFalcon & 0.76 & 0.14 & 0.04 & 0.00 & $1.4 \times 10^6$ \\
MKEvolve (PostFuse) & \textbf{\underline{0.94}} & 0.74 & 0.34 & 0.08 & $2.4 \times 10^6$ \\
MKEvolve (Swap) & \textbf{\underline{0.94}} & \textbf{\underline{0.80}} & \textbf{\underline{0.60}} & 0.06 & $2.1 \times 10^6$ \\
\bottomrule
\end{tabular}%
}
\label{tab:c4o_l3}
\end{subtable}

\vspace{1em}

\caption{
Claude 4.5 Opus experiment results on baseline correctness and speedup. For each metric, the best value among the first four baselines synthesizing 160 kernels is shown in bold. The best value across all methods is shown in bold and underlined.}
\label{tab:c4o_main}
\vspace{-1em}
\end{table*}

\paragraph{Additional Remarks} 
\tsc{MKEvolve} produces interpretable and modular kernels that enable engineers to quickly localize which subproblem is responsible for end-to-end errors (see \Cref{fig:mainfigure}), rather than relying solely on aggregate failure messages that only report overall numerical deviation of the kernel output from the ground truth reference implementation.
\tsc{MKEvolve} also enables flexible selection among subproblem kernels and direct observation of each subkernel’s impact on overall speedup (see \Cref{fig:mainfigure}).
Notably, even without invoking the LLMFuse agent, \tsc{MKEvolve} supports kernel fusion–based optimizations at the subproblem level.
Finally, under certain assumptions, the global model error can be bounded by the local errors of individual subkernels, ensuring correctness of both the overall kernel and its components (see \Cref{appendix:localglobalproof}).


\section{Experiments}

\paragraph{Setup} We assess the correctness and speedup of LLM-synthesized FP32 Triton kernels on 150 KernelBench \citep{ouyang2025kernelbench} L2 and L3 benchmark problems.
The KernelBench L2 benchmark comprises 100 PyTorch problems that include multiple primitive operator sequences (e.g., a combination of convolution, ReLU, and bias).
The KernelBench L3 benchmark comprises 50 PyTorch problems that include full machine learning models (e.g., AlexNet, MiniGPT).
All experiments are conducted on AWS P4d nodes with A100 GPUs and are conducted twice using Claude 4.5 Opus and GPT-OSS 120B as the base LLM.

We assess a baseline’s performance using the following metrics:
(1) the proportion of benchmark problems for which it generates a correct solution (Correct),
(2) the proportion of which it produces a kernel that is at least $p$-times as fast as \textit{torch.compile} (Fast$_p$ where $p \in \{0.5, 1, 2\}$) \citep{ouyang2025kernelbench}, and 
(3) the proportion of which it produces a kernel that is faster than all other baselines.
These metrics are computed after the baseline finishes and returns the fastest discovered kernel.
All kernels are evaluated with the KernelBench repository's CUDA-stream evaluation pipeline.
The pipeline measures correctness by running each kernel 5 times with a random input of same shape and comparing the outputs to the PyTorch output with atol/rtol of 10$^{-4}$.
Speedup is measured against \textit{torch.compile} using 10 warmup and 100 profiling runs, with correctness check failure defined as 0 speedup.
We also augment our evaluation pipeline with a TritonRL-based \citep{woo2025tritonrl} cheating detector and impose a 5-minute subprocess-level time limit. 

We benchmark Parallel Scaling, Beam Search, MKEvolve, and MKEvolve (Fuse) that generate 160 kernels for each problem with the same set of prompts and evaluators.
This isolates the effects of subkernel optimization vs end-to-end kernel optimization on the solution correctness and speedup.
We also benchmark additional baselines that generate a variable number of kernels:
KernelFalcon \citep{wang2025kernelfalcon} \footnote{We baseline the March 3rd, 2026 version of KernelFalcon.} modified to generate FP32 kernels that are evaluated with our pipeline, MKEvolve (PostFuse) that attempts to fuse MKEvolve's solution into an end-to-end kernel 24 times, and MKEvolve (Swap), which programmatically replaces MKEvolve's solution subkernels with less than 0.9x speedup of subproblem \textit{torch.compile} code with PyTorch implementations.
Please refer to Appendix \ref{appendix:expsetup} for additional details.

\paragraph{Claude 4.5 Opus Results}

\begin{table*}[t]
\centering
\small

\begin{subtable}{0.775\linewidth}
\centering
\caption{GPT-OSS 120B metrics on 100 KernelBench level 2 problems.}
\begin{tabular}{lccccc}
\toprule
\textsc{Method} & Correct (↑) & Fast$_{0.5}$ (↑) & Fast$_{1}$ (↑) & Fast$_{2}$ (↑) & \# Tokens \\
\midrule
Parallel Scaling & 0.59 & 0.32 & 0.13 & 0.05 & $1.6 \times 10^6$ \\
Beam Search & 0.84 & 0.48 & 0.23 & \textbf{\underline{0.16}} & $2.5 \times 10^6$ \\
MKEvolve & 0.94 & \textbf{0.54} & 0.22 & 0.05 & $1.9 \times 10^6$ \\
MKEvolve (Fuse) & \textbf{0.95} & 0.53 & \textbf{0.24} & 0.15 & $2.2 \times 10^6$ \\
\midrule
KernelFalcon & 0.83 & 0.25 & 0.01 & 0.00 & $0.3 \times 10^6$ \\
MKEvolve (PostFuse) & 0.97 & 0.54 & 0.27 & 0.08 & $2.3 \times 10^6$ \\
MKEvolve (Swap) & \textbf{\underline{0.99}} & \textbf{\underline{0.77}} & \textbf{\underline{0.35}} & 0.05 & $1.9 \times 10^6$ \\
\bottomrule
\end{tabular}%
\label{tab:oss_l2}
\end{subtable}

\vspace{0.75em}

\begin{subtable}{0.775\linewidth}
\centering
\caption{GPT-OSS 120B metrics on 50 KernelBench level 3 problems.}
\begin{tabular}{lccccc}
\toprule
\textsc{Method} & Correct (↑) & Fast$_{0.5}$ (↑) & Fast$_{1}$ (↑) & Fast$_{2}$ (↑) & \# Tokens \\
\midrule
Parallel Scaling & 0.26 & 0.12 & 0.02 & 0.00 & $2.0 \times 10^6$ \\
Beam Search & 0.52 & 0.14 & 0.04 & \textbf{\underline{0.02}} & $3.8 \times 10^6$ \\
MKEvolve & 0.70 & 0.18 & 0.06 & \textbf{\underline{0.02}} & $2.4 \times 10^6$ \\
MKEvolve (Fuse) & \textbf{0.72} & \textbf{0.32} & \textbf{0.12} & 0.00 & $2.7 \times 10^6$ \\
\midrule
KernelFalcon & 0.38 & 0.10 & 0.02 & 0.00 & $1.0 \times 10^6$ \\
MKEvolve (PostFuse) & 0.72 & 0.20 & 0.10 & \textbf{\underline{0.02}} & $3.0 \times 10^6$ \\
MKEvolve (Swap) & \textbf{\underline{0.88}} & \textbf{\underline{0.58}} & \textbf{\underline{0.36}} & 0.02 & $2.4 \times 10^6$ \\
\bottomrule
\end{tabular}%
\label{tab:oss_l3}
\end{subtable}
\vspace{1em}
\caption{
GPT-OSS 120B experiment results on baseline correctness and speedup. For each metric, the best value among the first four baselines synthesizing 160 kernels is shown in bold. The best value across all methods is shown in bold and underlined.}
\label{tab:oss_main}
\vspace{-1em}
\end{table*}

\begin{figure*}[t]
    \centering
    \includegraphics[width=0.75\linewidth]{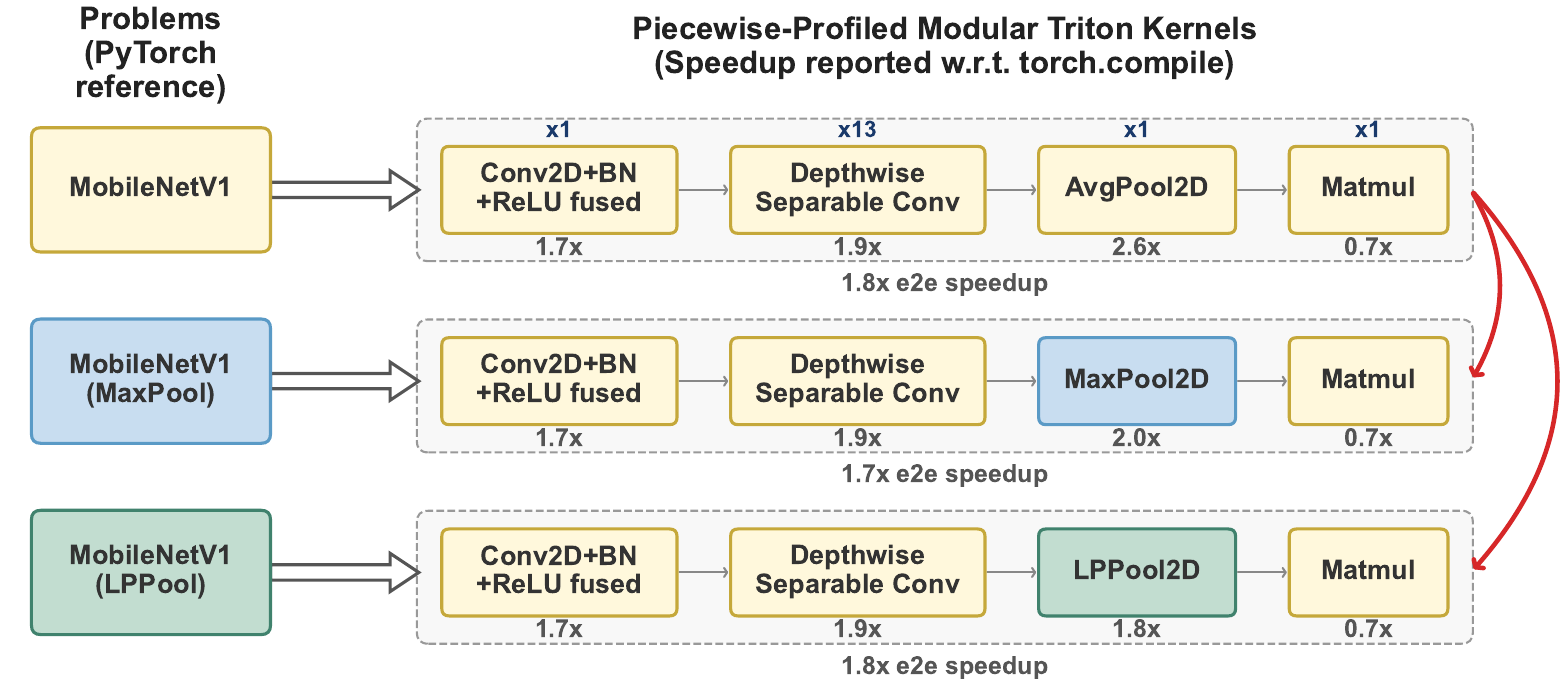}
    \captionsetup{font=small}
    \caption{\small 
    Transferability of MKEvolve-generated Triton kernels across MobileNetV1 variants. {\bf Top}: The base kernel invokes four subkernels sequentially, achieving 1.8× speedup over torch.compile. {\bf Middle \& Bottom}: Replacing AvgPool2D with MaxPool2D or LPPool2D yields adapted kernels achieving 1.7× and 1.8× speedups, requiring only 4 LLM calls to produce the new pooling subkernels.
    }
    \label{fig:adaptation}
\end{figure*}

\begin{figure}[t]
    \centering
    \includegraphics[width=\linewidth]{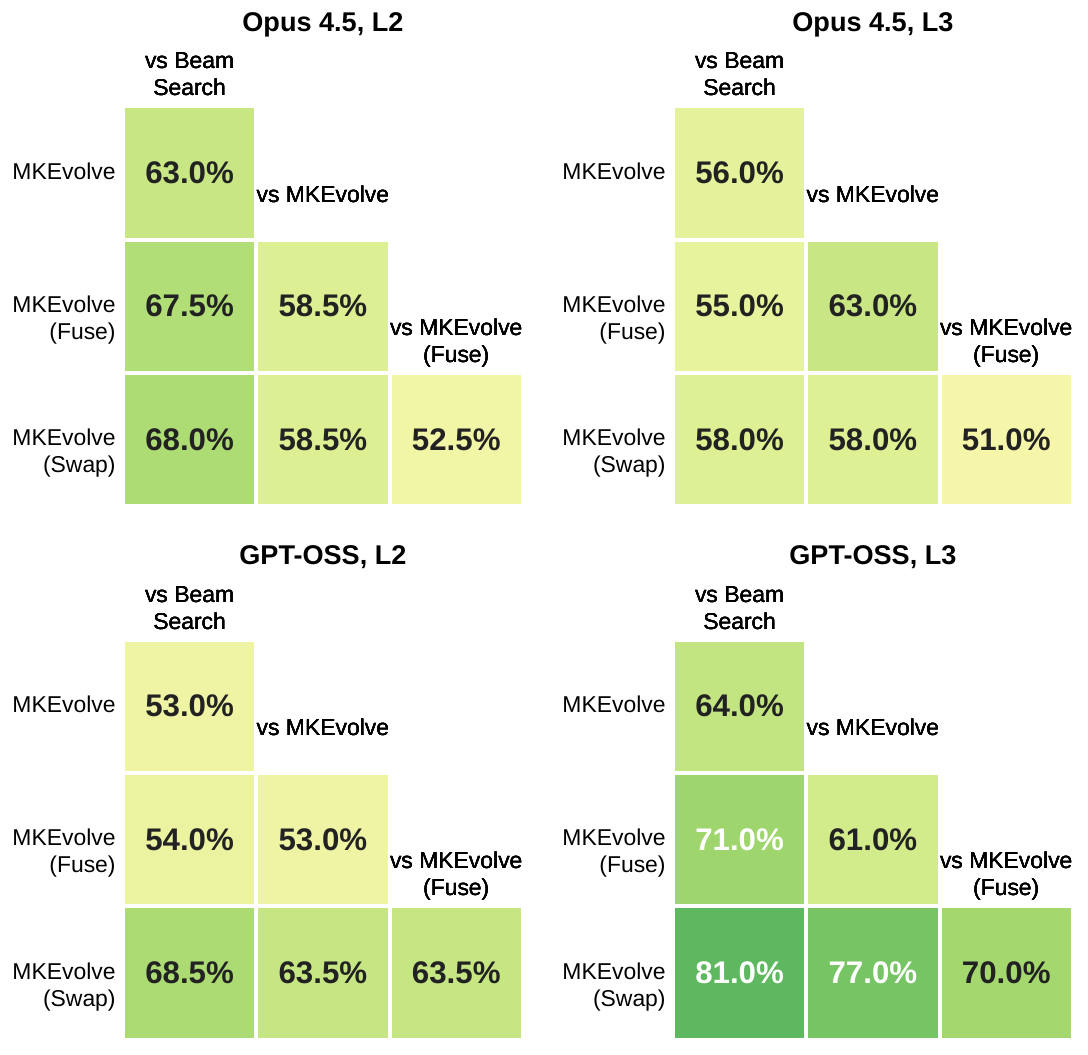}
    \captionsetup{font=small}
    \caption{\small 
    Win-rate (proportion of tasks for which the row method achieved higher speedup than the column method) heatmaps comparing various MKEvolve variants and beam search, using both Claude 4.5 Opus and gpt-oss as backbone LLM to solve L2 and L3 tasks sampled from KernelBench.
    }
    \label{fig:heatmaps}
\end{figure}

\Cref{tab:c4o_main} reports correctness and speedup metrics for each baseline, along with the number of LLM tokens used during program execution.
We observe that MKEvolve outperforms Parallel Scaling and Beam Search on all but one metric across both KernelBench L2 and L3 benchmarks.
This result is achieved while using 15 to 30\% less tokens than beam search in spite of MKEvolve employing beam search under the hood.
We attribute this token efficiency to MKEvolve refining subkernels rather than optimizing end-to-end kernels, as done by the other baselines.
MKEvolve (Fuse) performs comparably to MKEvolve for the most part, though its attempt at fusing across subproblems and working with larger subproblems brings its token usage closer to Beam Search.
We note that MKEvolve is capable of producing fully fused kernels at the subproblem level.

When expanding the comparison to methods that synthesize a variable number of kernels (KernelFalcon) or build on MKEvolve's solutions, we find that MKEvolve (Swap) dominates most metrics.
While this success originates from selectively swapping subproblem solutions with PyTorch implementations, we note that the swapping behavior is entirely controllable and can be useful in practice when writing kernels for complex models, for which it is easy to design performant kernels for some parts but not others.
We find that KernelFalcon’s emphasis on synthesizing correct kernels, combined with its use of an LLM-determined correctness threshold and the absence of cheating detectors in parts of the codebase, could result in suboptimal performance.
The synthesized kernels were generally slow and sometimes failed our evaluation pipeline's correctness checks from exceeding the evaluation timeout, failing stricter correctness thresholds, or being flagged by the cheating detector.

Our findings are further supported by \Cref{fig:heatmaps}, which shows that MKEvolve and its variants produce faster kernels more frequently than Beam Search's kernels across benchmarks.
Overall, MKEvolve, through its modular subkernel synthesis strategy, outperforms the other baselines while using fewer LLM tokens on both the simpler operator-sequence tasks in KernelBench L2 and the more complex full-model tasks in KernelBench L3.

\paragraph{GPT-OSS 120B Results} \Cref{tab:oss_main} reports correctness and speedup metrics for each baseline, along with the number of LLM tokens used during program execution.
We observe a similar trend to that of the Claude 4.5 Opus results, where MKEvolve generally outperforms Beam Search on correctness and performance while using 25 to 35\% fewer tokens.
While the speedup metrics are closer compared to the Claude 4.5 Opus results, \Cref{fig:heatmaps} suggests that MKEvolve still holds an edge over Beam Search while being much more token efficient from solving simpler subproblems when synthesizing new kernels.
MKEvolve (Fuse) performs marginally better than MKEvolve at the expense of higher token usage.
MKEvolve (Swap) again performs best out of all baselines.
Overall, we find that the benefit of MKEvolve's modular kernel synthesis strategy can hold for both frontier and open-source base LLMs.

\paragraph{Further Analysis}

\Cref{fig:l2-c4o-timeline} and \Cref{fig:l3-c4o-timeline} in Appendix~\ref{appendix:timelines} depict the baselines' metric improvements over time as they generate 160 kernels versus tokens usage.
We observe that MKEvolve achieves the highest metrics across different token usage levels for most metrics across benchmarks.
We also ablate MKEvolve to perform a single problem decomposition in Appendix \ref{appendix:singlesplit}, and find that this slightly hurts the correctness metric compared to multiple subproblem splits.

We also rerun our main experiment for Beam Search and MKEvolve for another seed in Appendix \ref{appendix:secondseed} to assess the effect of random seed on our results.
We observe that MKEvolve generally outperforms Beam Search while using substantially fewer tokens, and on the Claude 4.5 Opus experiments, outperforms Beam Search on every metric and benchmark.
In addition, Appendix~\ref{appendix:llmkernels} reports a small-scale evaluation on three LLM inference kernels.

We emphasize that MKEvolve’s ability to operate on a decomposed problem representation enables it to generate kernels that are readily adaptable to related tasks, a desirable property in machine learning where practitioners often work with models that differ only slightly.
\Cref{fig:adaptation} showcases this for the KernelBench L3 MobileNetV1 problem.
By replacing the MobileNetV1 solution's AvgPool2D subkernel with MaxPool2D and LPPool2D subkernels synthesized after just 4 LLM invocations, MKEvolve can rapidly produce kernels that obtain 1.7x and 1.8x speedup over \textit{torch.compile} for the MobileNetV1 architectural variants.

Lastly, while we employed atol/rtol of $10^{-4}$ for the FP32 kernel correctness verification to be consistent with the official KernelBench evaluation, we empirically show how modular kernel generation with local subkernel verification paves the way for a more robust correctness check mechanism capable of detecting common correctness errors in Appendix \ref{appendix:tolerance}.

\section{Conclusion}

We introduced \tsc{MKEvolve}, a modular kernel code generation framework that iteratively decomposes complex PyTorch modules into subproblems, independently solves them, and composes the resulting subkernels into a complete, end-to-end solution.
We demonstrated that MKEvolve improves both kernel correctness and performance speedup, while reducing LLM token usage by up to 35\% compared to a beam search baseline that operates directly on end-to-end kernels.
We conclude by highlighting several promising directions for future work, including adaptive correctness threshold selection for subproblems, investigating more advanced subkernel optimization strategies such as Monte Carlo Tree Search, and extending \tsc{MKEvolve} to additional kernel programming frameworks and settings.

\section*{Impact Statement}

This work has the potential to accelerate AI training and inference on GPUs and NPUs.
Such improvements could enable more efficient AI systems, including systems that process and respond to user queries more quickly.
At the same time, the methods developed in this work could also be used in applications with negative societal consequences, such as more efficient surveillance systems.
As with many advances in AI infrastructure, the broader impact depends on how these methods are deployed.

\bibliography{example_paper}
\bibliographystyle{icml2026}

\newpage
\appendix
\onecolumn

\section{Additional Experiment Details}
\label{appendix:expsetup}

\paragraph{Shared Setup} 
All methods share the initial kernel generation prompt. Beam Search and MKEvolve variants use a beam width of 4, a beam expansion size of 1, and share the kernel debugging and refinement prompts.
MKEvolve updates its subgraph structure once every 2 iterations, runs for 5 outer loops with a per-iteration LLM call budget of 32, and does not use fusion.
We disable TF32 for both PyTorch and Triton as it affected correctness results for the default KernelBench threshold and ensure that the model tensor is contiguous.
For parallel scaling and beam search, only the solutions that pass the cheating detector one more time at the end are returned.
For MKEvolve, only subkernels that pass the cheating detector one more time after subkernel beam search are accepted.
Claude 4.5 Opus experiments employed Claude 4.5 Sonnet as the cheating detector model, whereas GPT-OSS 120B experiment employed GPT-OSS 120B as the cheating detector model.
We note that the typical wall clock time was around 2-4 hours per job.

\paragraph{KernelFalcon Setup} We baseline the March 3rd, 2026 version of the official KernelFalcon repository.
To match our evaluation setup, we changed KernelFalcon's self-test generation prompt to test FP32 kernels from FP16 kernels and set the default torch.allclose threshold default targets to 1e-4 atol/rtol for all problems (we note that KernelFalcon LLM-generated self-test allows the LLM to change the atol/rtol threshold).
KernelFalcon was informed that it is to design kernels that will run on A100 GPU and that tl.dot's allow\_tf32 flag can be set to False if there are persistent correctness errors.
We evaluated KernelFalcon with the same code that we used to evaluate all other approaches.
Because KernelFalcon's LLM-generated self-test outputs code is incompatible with KernelBench's evaluation script, we use Claude 4.5 Opus to convert the KernelFalcon solution wrapper to KernelBench evaluation-compatible format.
Specifically, we attempted conversion 5 times per kernel and considered the task as a success if any of the 5 conversion attempts succeeded.
We note that we further robustified the KernelFalcon repository to improve its reliability.
For example, we added retry logic to the subproblem structure JSON file creation that fixed KernelFalcon from aborting the subproblem decomposition attempt and attempting end-to-end sequential refinements.
In addition, we ensured that only a single subproblem kernel refinement happens at a time, since in some instances, KernelFalcon's AutoAgent attempted to launch 47 workers in parallel that attempt to evaluate 47 different subproblems on a single GPU, which led to GPU memory errors.
KernelFalcon was run with sequential scaling \lstinline[language={},keywordstyle=]{--max-iters=10} and \lstinline[language={},keywordstyle=]{--compose-max-iters=10}, which corresponds to a maximum of 40 refinement iterations for every subproblem kernel synthesis, 10 refinement iterations for LLM that combines the subkernels by including them in the prompt, and 40 refinement iterations for the backup sequential scaling agent that tries to build an end-to-end kernel from scratch.
KernelFalcon received the same verifier and profiler feedback from our evaluation pipeline as all other baselines.

\section{Timeline Plots}
\label{appendix:timelines}

\begin{figure}[h!]
    \centering
    \includegraphics[width=0.75\linewidth]{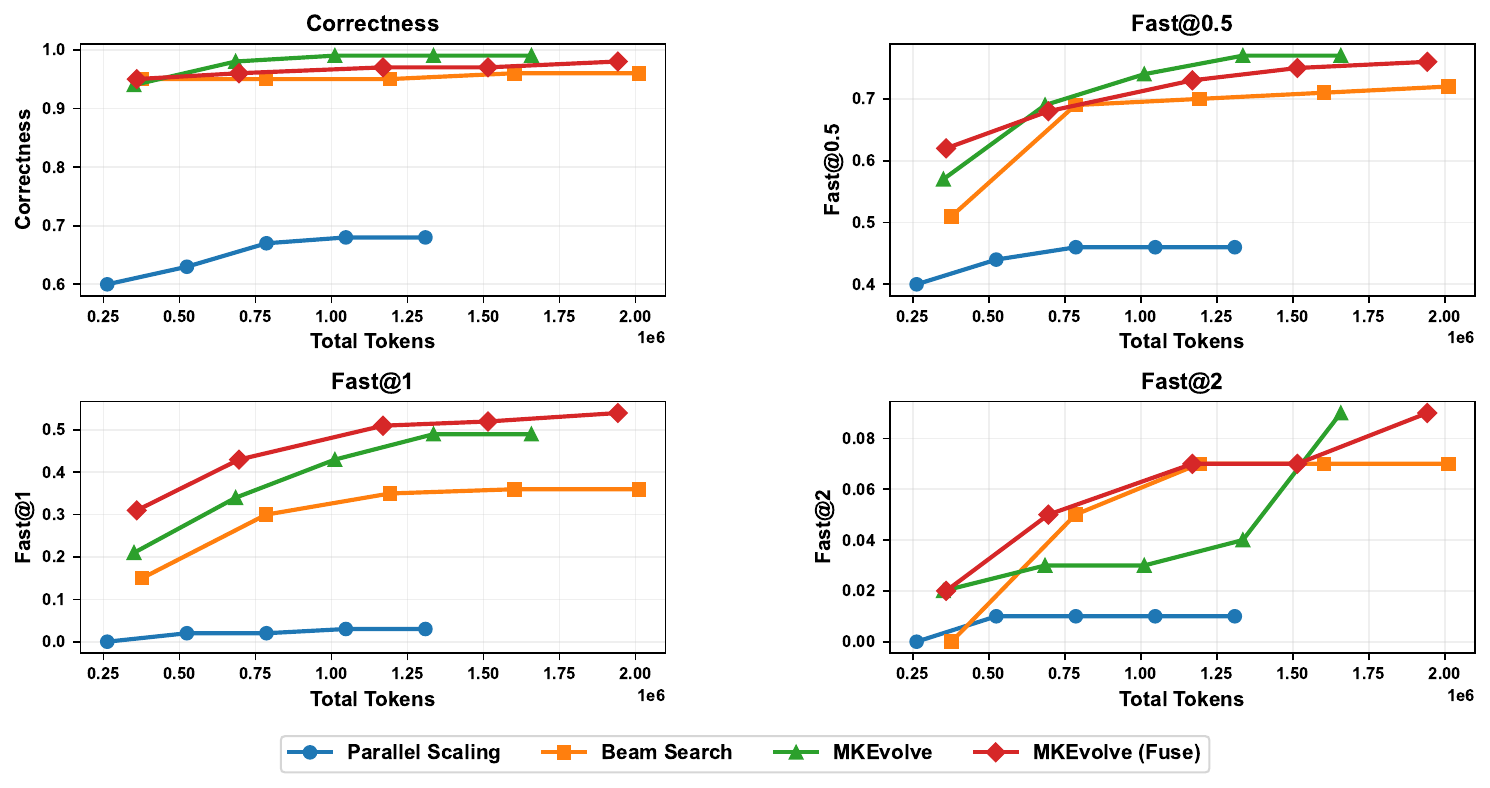}
    \captionsetup{font=small}
    \vspace*{-0.25em}
    \caption{\small Claude 4.5 Opus KernelBench L2 experiment improvement over time plot for all baselines. We note that MKEvolve and its variants achieve the best final performance on all metrics while consuming fewer tokens per outer iteration.}
    \label{fig:l2-c4o-timeline}
\end{figure}

\begin{figure}[h!]
    \centering
    \includegraphics[width=0.75\linewidth]{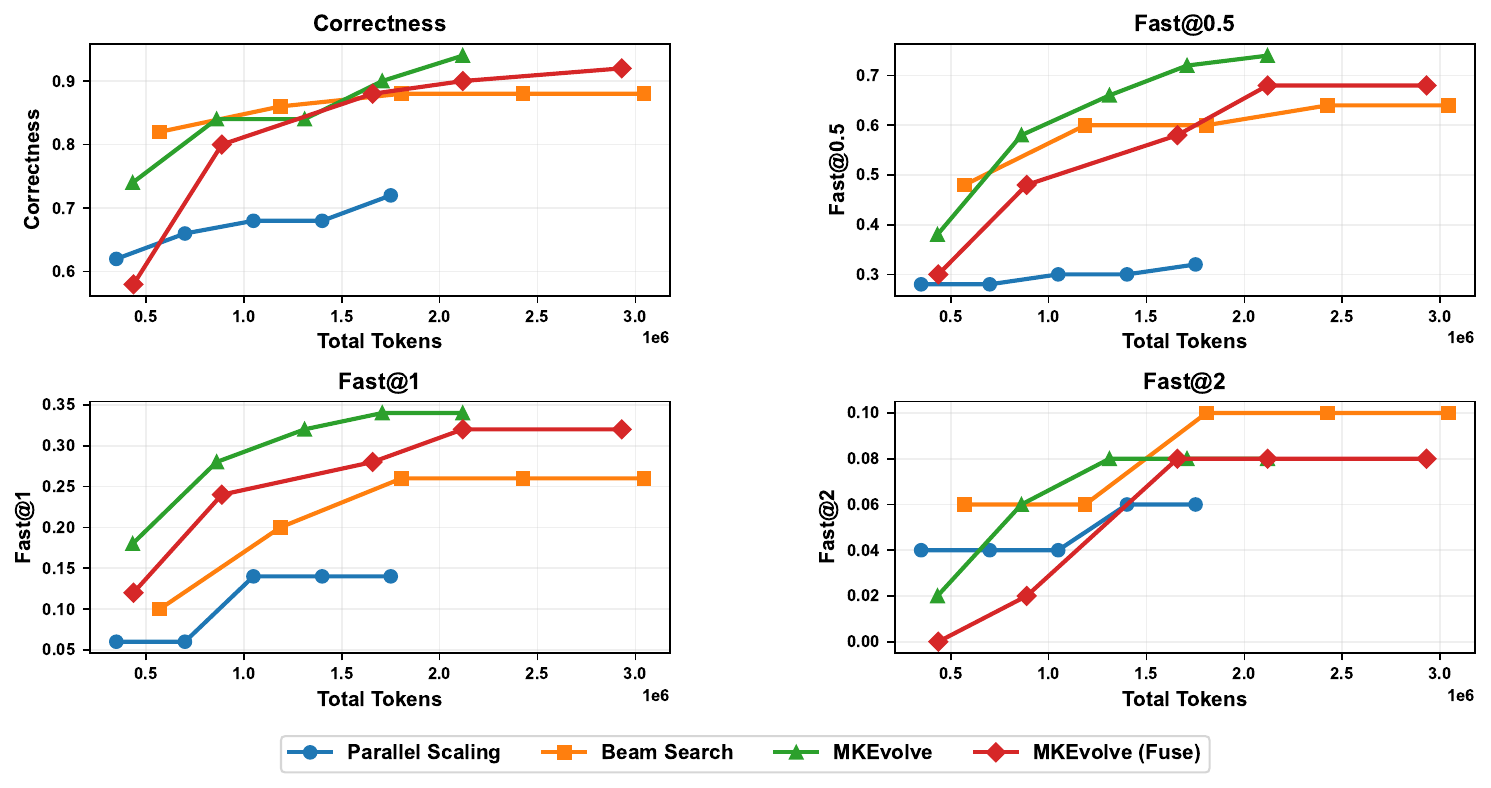}
    \captionsetup{font=small}
    \vspace*{-0.25em}
    \caption{\small Claude 4.5 Opus KernelBench L3 experiment improvement over time plot for all baselines. We note that MKEvolve and its variants achieve the best final performance on all but one metric while consuming fewer tokens per outer iteration.}
    \label{fig:l3-c4o-timeline}
    \vspace*{-0.25em}
\end{figure}

\section{Example Second Seed Results}
\label{appendix:secondseed}

\begin{table}[H]
\centering
\begin{tabular}{lcccccc}
\toprule
\textsc{Method} & Correct (↑) & Fast$_{0.5}$ (↑) & Fast$_{1}$ (↑) & Fast$_{2}$ (↑) & \# Tokens \\
\midrule
Beam Search & 0.95 & 0.66 & 0.37 & 0.03 & $2.0 \times 10^6$ \\
MKEvolve & \textbf{0.99} & \textbf{0.75} & \textbf{0.49} & \textbf{0.04} & $1.7 \times 10^6$ \\

\midrule
MKEvolve (Swap) & 1.00 & 0.87 & 0.61 & 0.04 & $1.7 \times 10^6$ \\

\bottomrule
\end{tabular}
\caption{Claude 4.5 Opus Metrics on 100 KernelBench level 2 problems (Second Seed)}
\end{table}

\vspace*{-0.25em}

\begin{table}[H]
\centering
\begin{tabular}{lcccccc}
\toprule
\textsc{Method} & Correct (↑) & Fast$_{0.5}$ (↑) & Fast$_{1}$ (↑) & Fast$_{2}$ (↑) & \# Tokens \\
\midrule
Beam Search & 0.88 & 0.64 & 0.26 & 0.06 & $3.0 \times 10^6$ \\
MKEvolve & \textbf{0.94} & \textbf{0.70} & \textbf{0.42} & \textbf{0.08} & $2.1 \times 10^6$ \\

\midrule
MKEvolve (Swap) & 0.96 & 0.78 & 0.58 & 0.04 & $2.1 \times 10^6$ \\

\bottomrule
\end{tabular}
\caption{Claude 4.5 Opus Metrics on 50 KernelBench level 3 problems (Second Seed)}
\end{table}

\vspace*{-0.25em}

\begin{table}[H]
\centering
\begin{tabular}{lcccccc}
\toprule
\textsc{Method} & Correct (↑) & Fast$_{0.5}$ (↑) & Fast$_{1}$ (↑) & Fast$_{2}$ (↑) & \# Tokens \\
\midrule
Beam Search & 0.87 & 0.47 & \textbf{0.29} & \textbf{0.17} & $2.5 \times 10^6$ \\
MKEvolve & \textbf{0.97} & \textbf{0.52} & 0.21 & 0.09 & $1.9 \times 10^6$ \\

\midrule
MKEvolve (Swap) & 0.99 & 0.76 & 0.27 & 0.08 & $1.9 \times 10^6$ \\

\bottomrule
\end{tabular}
\caption{GPT-OSS 120B Metrics on 100 KernelBench level 2 problems (Second Seed)}
\end{table}

\vspace*{-0.25em}

\begin{table}[H]
\centering
\begin{tabular}{lcccccc}
\toprule
\textsc{Method} & Correct (↑) & Fast$_{0.5}$ (↑) & Fast$_{1}$ (↑) & Fast$_{2}$ (↑) & \# Tokens \\
\midrule
Beam Search & 0.56 & \textbf{0.28} & 0.06 & \textbf{0.02} & $3.7 \times 10^6$ \\
MKEvolve & \textbf{0.68} & 0.24 & \textbf{0.10} & \textbf{0.02} & $2.4 \times 10^6$ \\

\midrule
MKEvolve (Swap) & 0.84 & 0.66 & 0.38 & 0.00 & $2.4 \times 10^6$ \\

\bottomrule
\end{tabular}
\caption{GPT-OSS 120B Metrics on 50 KernelBench level 3 problems (Second Seed)}
\end{table}

\section{Modular Kernel Generation Enables Better Correctness Checking}
\label{appendix:tolerance}

KernelBench's approach for verifying kernel correctness is to compare output tensors
against CPU references using \texttt{torch.allclose(output, reference, atol, rtol)} and pre-determined thresholds.
This function checks whether $|\text{output} - \text{reference}| \leq \texttt{atol} + \texttt{rtol} \times |\text{reference}|$
holds element-wise. While the criteria itself seems to be simple, we argue that choosing appropriate tolerance values is
fundamentally difficult in practice, especially for lower-precision beyond FP32. The tolerance needs to depend on various factors like datatype, operator, operand size, input data distribution, etc. For more complex kernels that involve multiple operators, this difficulty is further amplified. Through this section, we argue that our decomposition technique makes the correctness checking problem more tractable to ensure the correctness of LLM-generated kernel implementations in practice.

\begin{table}[t]
\centering
\caption{Maximum absolute error (GPU vs.\ FP32 CPU reference) across operations.
Errors collected using 30 runs with random $\mathcal{N}(0,1)$ inputs.}
\label{tab:error-profiles}
\small
\begin{tabular}{@{}lcccc@{}}
\toprule
 & \multicolumn{2}{c}{\textbf{GPU FP32}} & \multicolumn{2}{c}{\textbf{GPU BF16}} \\
\cmidrule(lr){2-3} \cmidrule(l){4-5}
\textbf{Operation} & \textbf{Max Abs} & \textbf{Mean Abs} & \textbf{Max Abs} & \textbf{Mean Abs} \\
\midrule
Matmul $512^2$              & \num{5e-5}--\num{7e-5}   & ${\sim}$\num{6e-6}
                            & \num{3e-1}--\num{5e-1}   & ${\sim}$\num{1e-5} \\
Softmax $128{\times}1024$   & \num{4e-9}--\num{7e-9}   & ${\sim}$\num{6e-11}
                            & \num{0}--\num{3e-5}       & ${\sim}$\num{2e-10} \\
LayerNorm                   & ${\sim}$\num{1e-6}        & ${\sim}$\num{4e-8}
                            & \num{8e-3}--\num{2e-2}    & ${\sim}$\num{5e-8} \\
CrossEntropy $128{\times}1000$ & ${\sim}$\num{1e-6}     & ${\sim}$\num{5e-8}
                            & \num{3e-2}--\num{3e-2}    & ${\sim}$\num{1e-2} \\
Conv2d                      & \num{8e-4}--\num{1e-3}    & ${\sim}$\num{1e-4}
                            & \num{1e-2}--\num{2e-2}    & ${\sim}$\num{9e-4} \\
\bottomrule
\end{tabular}
\end{table}

First, we demonstrate that the floating-point errors inherent to operators can make it complicated to determine the right threshold, even when considering operators in isolation.
Table~\ref{tab:error-profiles} shows the maximum absolute error between correct GPU and
FP32 CPU implementations: in FP32, errors range from
${\sim}\num{4e-9}$ (softmax) to ${\sim}\num{1e-3}$ (conv2d);
In BF16 from ${\sim}\num{3e-5}$ (softmax) to ${\sim}\num{5e-1}$
(matmul). Table~\ref{tab:input-dep} shows that the same matmul produces errors spanning
many orders of magnitude depending on input scale.
Table~\ref{tab:dim-scaling} shows error also grows with matrix dimension.
These observations offer some intuition for why checking correctness of complex kernels using the resulting tensors can be difficult: the inherent ``error profile'' for different operators can differ dramatically, and can even depend on the output of preceding operators; also numerical deviations caused by incorrect components may be ``partially absorbed'' by later components, due to their differences in sensitivity to numerical perturbations.

\begin{table}[t]
\centering
\caption{Input-dependent error for matmul $1024{\times}1024$ (max absolute error, 30 runs).}
\label{tab:input-dep}
\small
\begin{tabular}{@{}lcc@{}}
\toprule
\textbf{Input Distribution} & \textbf{FP32 Max Abs} & \textbf{BF16 Max Abs} \\
\midrule
$\mathcal{N}(0, 0.01^2)$        & \num{2.0e-8}--\num{2.6e-8}     & \num{3.1e-5}--\num{6.1e-5} \\
$\mathcal{N}(0, 1)$              & \num{2.0e-4}--\num{2.7e-4}     & \num{4.7e-1}--\num{5.0e-1} \\
$\mathcal{N}(0, 10^2)$           & \num{1.9e-2}--\num{2.8e-2}     & \num{3.2e+1}--\num{6.1e+1} \\
$\text{Uniform}(0, 1)$           & \num{5.2e-4}--\num{6.7e-4}     & ${\sim}$\num{1.0} \\
$\text{Uniform}(-100, 100)$      & \num{6.3e-1}--\num{8.8e-1}     & \num{1.0e+3}--\num{2.0e+3} \\
Sparse (5\%)                      & \num{1.4e-6}--\num{1.9e-6}     & \num{3.1e-2}--\num{6.0e-2} \\
\bottomrule
\end{tabular}
\end{table}

\begin{table}[t]
\centering
\caption{Error growth with matrix dimension for matmul ($\mathcal{N}(0,1)$ inputs, 30 runs).
BF16 errors are ${\sim}\num{1000}{\times}$ larger than FP32 and both grow with dimension.}
\label{tab:dim-scaling}
\small
\begin{tabular}{@{}rcc|cc@{}}
\toprule
 & \multicolumn{2}{c|}{\textbf{FP32}} & \multicolumn{2}{c}{\textbf{BF16}} \\
\cmidrule(lr){2-3} \cmidrule(l){4-5}
\textbf{$N$} & \textbf{Avg Max Abs} & \textbf{Avg Mean Abs} & \textbf{Avg Max Abs} & \textbf{Avg Mean Abs} \\
\midrule
$64$    & \num{0}       & \num{0}       & \num{6.9e-2}  & \num{9.0e-3} \\
$128$   & \num{0}       & \num{0}       & \num{1.2e-1}  & \num{1.3e-2} \\
$256$   & \num{4.6e-5}      & \num{3.6e-6}      & \num{2.0e-1}  & \num{1.8e-2} \\
$512$   & \num{6.1e-5}      & \num{5.8e-6}      & \num{2.5e-1}  & \num{2.5e-2} \\
$1024$  & \num{2.1e-4}      & \num{1.3e-5}      & \num{4.9e-1}  & \num{3.6e-2} \\
$2048$  & \num{5.2e-4}      & \num{2.7e-5}      & \num{5.0e-1}  & \num{5.1e-2} \\
$4096$  & \num{1.0e-3}      & \num{5.4e-5}      & \num{1.0}     & \num{7.2e-2} \\
\bottomrule
\end{tabular}
\end{table}

We now show that decomposing composed operators into sub-operations makes per-operator
tolerance selection more reliable through case studies.
We use two realistic cases:
(1)~MinGPT Causal Attention (KernelBench \texttt{L3\#43}: QKV projection $\to$ reshape $\to$
scaled dot-product $\to$ causal mask $\to$ softmax $\to$ attention $\to$ output projection),
and (2)~a Transformer FFN block (LayerNorm $\to$ Linear $\to$ GELU $\to$ Linear $\to$ residual).
We inject bugs that reflect common mistakes when writing Triton kernels, which include the following
attention bugs:
\begin{itemize}
\item \textbf{Unsafe softmax:} \texttt{exp(x)} without subtracting the max---overflows for large scores.
\item \textbf{Causal mask off-by-one:} diagonal is masked (\texttt{tril(diagonal=-1)} vs \texttt{tril(diagonal=0)}).
\item \textbf{Tile boundary ($K$):} last partial tile of $K$'s head dimension is zeroed (unloaded), simulating missing boundary handling when dimensions don't divide evenly by \texttt{BLOCK\_SIZE}.
\item \textbf{Wrong accumulator dtype:} $QK^\top$ matmul accumulates in BF16 instead of FP32.
\item \textbf{Softmax wrong dimension:} softmax computed over \texttt{dim=-2} instead of \texttt{dim=-1}.
\end{itemize}
and FFN bugs:
\begin{itemize}
\item \textbf{Unsafe LayerNorm:} divides by $\sigma$ without subtracting $\mu$ first.
\item \textbf{GELU sigmoid approximation:} uses $x \cdot \sigma(1.702 x)$ instead of the exact GELU.
\item \textbf{Tile boundary (fc1):} last partial block of \texttt{fc1.weight} is zeroed (unloaded).
\item \textbf{Wrong accumulator dtype (fc2):} fc2 matmul accumulates in BF16 instead of FP32.
\item \textbf{Missing residual connection:} output $= \text{fc2}(\text{GELU}(\text{fc1}(\text{LN}(x))))$ without adding $x$.
\item \textbf{ReLU instead of GELU:} wrong activation function.
\end{itemize}

We compare two verification strategies. \textbf{End-to-end (E2E):} check only the final
output using a fixed tolerance, as in KernelBench.
\textbf{Decomposed:} check each intermediate tensor using a calibrated per-operator tolerance
($\text{max} + 3\sigma$ of the correct error observed over 20 calibration runs, with
$\texttt{rtol}{=}0$); a bug is caught if any stage fails.
Tables~\ref{tab:bf16-dilemma-attn} and~\ref{tab:bf16-dilemma-ffn} show the core problem
with fixed thresholds for BF16 composed operators: every choice is either too tight
(rejecting correct outputs) or too loose (missing bugs). For the attention block,
$\texttt{atol}=10^{-3}$ rejects 100\% of correct outputs; $\texttt{atol}=10^{-2}$ is
usable but misses 2 of 5 bugs. For the FFN block, $\texttt{atol}=10^{-2}$ rejects 100\%
of correct outputs, while $\texttt{atol}=10^{-1}$---the only usable option---misses
2 of 6 bugs.

\begin{table}[t]
\centering
\caption{BF16 threshold dilemma for \textbf{Causal Attention}: false negative rate
(correct outputs rejected) and bug catch rate at three fixed E2E thresholds (out of 20 eval runs).}
\label{tab:bf16-dilemma-attn}
\small
\begin{tabular}{@{}lccc@{}}
\toprule
 & $\texttt{atol}{=}10^{-3}$ & $\texttt{atol}{=}10^{-2}$ & $\texttt{atol}{=}10^{-1}$ \\
\midrule
Correct rejected       & \textbf{20/20} & 0/20 & 0/20 \\
\midrule
Unsafe softmax         & 20/20 & \textbf{0/20} & \textbf{0/20} \\
Mask off-by-one        & 20/20 & 20/20 & 20/20 \\
Tile boundary ($K$)    & 20/20 & 20/20 & 20/20 \\
Wrong acc dtype ($QK$) & 20/20 & \textbf{0/20} & \textbf{0/20} \\
Softmax dim=$-2$       & 20/20 & 20/20 & 20/20 \\
\bottomrule
\end{tabular}
\end{table}

\begin{table}[t]
\centering
\caption{BF16 threshold dilemma for \textbf{FFN block}: false negative rate and bug catch rate
at three fixed E2E thresholds (out of 20 eval runs).}
\label{tab:bf16-dilemma-ffn}
\small
\begin{tabular}{@{}lccc@{}}
\toprule
 & $\texttt{atol}{=}10^{-3}$ & $\texttt{atol}{=}10^{-2}$ & $\texttt{atol}{=}10^{-1}$ \\
\midrule
Correct rejected          & \textbf{20/20} & \textbf{20/20} & 0/20 \\
\midrule
Unsafe LayerNorm          & 20/20 & 20/20 & \textbf{5/20} \\
GELU sigmoid approx       & 20/20 & 20/20 & \textbf{0/20} \\
Tile boundary (fc1)       & 20/20 & 20/20 & 20/20 \\
Wrong acc dtype (fc2)     & 20/20 & 20/20 & \textbf{0/20} \\
Missing residual          & 20/20 & 20/20 & 20/20 \\
ReLU not GELU             & 20/20 & 20/20 & 20/20 \\
\bottomrule
\end{tabular}
\end{table}

Tables~\ref{tab:detect-attn} and~\ref{tab:detect-ffn} compare E2E at the best usable
fixed threshold against the decomposed approach.
In FP32, the decomposed approach matches or exceeds E2E on every bug, but the gap is very small.
In BF16, the gap between the E2E and the decomposed approach becomes much more obvious. 
The decomposed method catches 4/5 attention bugs (vs.\ 3/5 for E2E and 5/6 FFN bugs (vs.\ 4/6 for E2E).

\begin{table}[t]
\centering
\caption{Bug detection for MinGPT Causal Attention (20 eval runs). E2E uses the best usable
fixed threshold (FP32: $10^{-5}$; BF16: $10^{-2}$). Decomposed uses max$+3\sigma$
per-stage tolerances with $\texttt{rtol}{=}0$. $^\dagger$Undetectable by both.}
\label{tab:detect-attn}
\small
\begin{tabular}{@{}lcc|cc@{}}
\toprule
 & \multicolumn{2}{c|}{\textbf{FP32}} & \multicolumn{2}{c}{\textbf{BF16}} \\
\cmidrule(lr){2-3} \cmidrule(l){4-5}
\textbf{Bug} & \textbf{E2E} & \textbf{Dec} & \textbf{E2E} & \textbf{Dec} \\
\midrule
Unsafe softmax         & 0$^\dagger$ & 1  & 0    & \textbf{12} \\
Mask off-by-one        & 20 & 20 & 20  & 20 \\
Tile boundary ($K$)    & 20 & 20 & 20  & 20 \\
Wrong acc dtype ($QK$) & 20 & 20 & 0$^\dagger$ & 0$^\dagger$ \\
Softmax dim=$-2$       & 20 & 20 & 20  & 20 \\
\midrule
\textbf{Total}    & 80/100 & \textbf{81/100} & 60/100 & \textbf{72/100} \\
Correct rejected  & 0/20 & 0/20 & 0/20 & 0/20 \\
\bottomrule
\end{tabular}
\end{table}

\begin{table}[t]
\centering
\caption{Bug detection for Transformer FFN block (20 eval runs). E2E uses fixed threshold
(FP32: $10^{-5}$; BF16: $10^{-1}$, the only usable option). $^\dagger$Undetectable by both.}
\label{tab:detect-ffn}
\small
\begin{tabular}{@{}lcc|cc@{}}
\toprule
 & \multicolumn{2}{c|}{\textbf{FP32}} & \multicolumn{2}{c}{\textbf{BF16}} \\
\cmidrule(lr){2-3} \cmidrule(l){4-5}
\textbf{Bug} & \textbf{E2E} & \textbf{Dec} & \textbf{E2E} & \textbf{Dec} \\
\midrule
Unsafe LayerNorm       & 20 & 20 & 5   & \textbf{20} \\
GELU sigmoid approx    & 20 & 20 & 0   & \textbf{20} \\
Tile boundary (fc1)    & 20 & 20 & 20  & 20 \\
Wrong acc dtype (fc2)  & 20 & 20 & 0$^\dagger$ & 0$^\dagger$ \\
Missing residual       & 20 & 20 & 20  & 20 \\
ReLU not GELU          & 20 & 20 & 20  & 20 \\
\midrule
\textbf{Total}    & 120/120 & \textbf{120/120} & 65/120 & \textbf{100/120} \\
Correct rejected  & 0/20 & 1/20 & 0/20 & 0/20 \\
\bottomrule
\end{tabular}
\end{table}

We empirically define the \emph{feasibility gap} of an operator $s$ to quantify the robustness of detection:
\begin{equation*}
\text{gap}(s) = \min_{\text{trials}} \text{err}_{\text{buggy}}(s) \;-\; \max_{\text{trials}} \text{err}_{\text{correct}}(s)
\label{eq:gap}
\end{equation*}
A positive gap means a tolerance exists that perfectly separates correct from buggy at the output of
operator $s$; a larger gap (bigger when both gaps are positive, or closer to zero when both gaps are negative) 
means the detection is more robust to input randomness
and hardware nondeterminism. A negative gap means no tolerance can separate correct from buggy.
Table~\ref{tab:feas-gap} shows that for bugs caught by both methods, the decomposed
feasibility gap is consistently wider.
For particular bugs like GELU sigmoid approximation and unsafe softmax in BF16, the E2E gap is
\textbf{negative}, while the gap is positive for the decomposed method.
This illustrates that the decomposed method can enable detection of certain subtle bugs that are 
impossible to be caught using the E2E checking method.

\begin{table}[t]
\centering
\caption{Feasibility gap: E2E (best usable fixed threshold) vs.\ decomposed (best stage).
A positive gap means the bug is separable from correct outputs; negative means inseparable.
``Ratio'' = decomposed gap / E2E gap;
$\infty$: E2E gap negative, decomposed gap positive.}
\label{tab:feas-gap}
\scriptsize
\begin{tabular}{@{}lrrc|rrc@{}}
\toprule
 & \multicolumn{3}{c|}{\textbf{FP32}} & \multicolumn{3}{c}{\textbf{BF16}} \\
\cmidrule(lr){2-4} \cmidrule(l){5-7}
\textbf{Bug} & \textbf{E2E} & \textbf{Dec (stage)} & \textbf{Ratio} & \textbf{E2E} & \textbf{Dec (stage)} & \textbf{Ratio} \\
\midrule
\multicolumn{7}{@{}l}{\emph{Causal Attention}} \\
\quad Unsafe softmax   & $-$\num{9e-7} & $-$\num{1e-7} (smax)  & n/a
                       & $-$\num{6e-4} & $+$\num{2e-4} (smax)  & $\boldsymbol{\infty}$ \\
\quad Mask off-by-one  & $+$\num{7.6e-1} & $+$\num{1.5} (attn)  & $2\times$
                       & $+$\num{7.7e-1} & $+$\num{1.5} (attn)  & $2\times$ \\
\quad Tile boundary    & $+$\num{1.1e-1} & $+$\num{11} ($QK$)   & $\mathbf{99\times}$
                       & $+$\num{1.1e-1} & $+$\num{11} ($QK$)   & $\mathbf{101\times}$ \\
\quad Wrong acc dtype  & $+$\num{5.6e-4} & $+$\num{6.3e-2} ($QK$) & $\mathbf{112\times}$
                       & $-$\num{1.1e-3} & $-$\num{1e-5} (proj) & n/a \\
\quad Softmax dim=$-2$ & $+$\num{1.4} & $+$\num{2.4} (attn) & $2\times$
                       & $+$\num{1.4} & $+$\num{2.5} (attn) & $2\times$ \\
\midrule
\multicolumn{7}{@{}l}{\emph{FFN Block}} \\
\quad Unsafe LN        & $+$\num{6.0e-2} & $+$\num{1.8e-1} (lin1)  & $3\times$
                       & $+$\num{5.9e-2} & $+$\num{2.6e-1} (lin1)  & $4\times$ \\
\quad GELU sigmoid     & $+$\num{9.9e-3} & $+$\num{2.0e-2} (gelu)  & $2\times$
                       & $-$\num{2.2e-3} & $+$\num{1.6e-2} (gelu)  & $\boldsymbol{\infty}$ \\
\quad Tile boundary    & $+$\num{1.0e-1} & $+$\num{2.1} (lin1)     & $\mathbf{21\times}$
                       & $+$\num{8.4e-2} & $+$\num{2.1} (lin1)     & $\mathbf{25\times}$ \\
\quad Wrong acc dtype  & $+$\num{3.1e-3} & $+$\num{3.1e-3} (lin2)  & $1\times$
                       & $-$\num{3.6e-3} & $-$\num{5.7e-4} (lin2)  & n/a \\
\quad Missing residual & $+$\num{4.3} & $+$\num{4.3} (res)      & $1\times$
                       & $+$\num{4.4} & $+$\num{4.4} (res)      & $1\times$ \\
\quad ReLU not GELU    & $+$\num{3.3e-1} & $+$\num{3.3e-1} (lin2) & $1\times$
                       & $+$\num{2.8e-1} & $+$\num{2.9e-1} (lin2) & $1\times$ \\
\bottomrule
\end{tabular}
\end{table}

In the current paper, we conduct all our correctness checking using the criteria of \texttt{atol=rtol=1e-4} for 
per-module checking and end-to-end result tensor checking, which agrees with what KernelBench provides, 
to ensure a fair comparison with existing techniques building on KernelBench.
Our correctness analysis above also supports this decision, due to the similar observed effects of using fine-grained accuracy criteria versus end-to-end criteria for FP32 kernels in KernelBench.
However, we point out that our modular kernel generation and accuracy checking framework paves the way for more robust accuracy checking
when it comes to generating kernels that use lower-precision datatypes in practice.

\section{Local Smoothness Implies Global Smoothness}
\label{appendix:localglobalproof}
There exist various work studying smoothness properties of neural network building blocks~\citep{kim2021lipschitzconstantselfattention,castin2024smoothattention,qi_understanding_2023} as well as techniques for estimating floating-
point round-off errors~\citep{beuzeville2025deterministic,higham_mary_2020}. Following standard procedures of probabilistic backward error analysis (see e.g., \citet{beuzeville2025deterministic}) on the $L_\infty$ norm of tensors, it is possible to get a bound on the element-wise maximum difference for the output tensor of a sequence of composed operators, which is expressed in terms of the condition number and other properties of the individual operators and their floating-point implementations. We note that this observation offers a guarantee that the tolerance for the composed operators can be guaranteed given the tolerance for individual operators, while the converse is not necessarily true---we demonstrated several such cases where numerical deviation are ``dampened'' by subsequent operators in \autoref{appendix:tolerance}. This offers some empirical evidence on why our compositional generation method is more robust in ensuring that all operators are implemented correctly.

\section{Single Split Results}
\label{appendix:singlesplit}

\begin{table}[H]
\centering

\begin{subtable}{\textwidth}\centering
\begin{tabular}{lcccccc}
\toprule
\textsc{Method} & Correct (↑) & Fast$_{0.5}$ (↑) & Fast$_{1}$ (↑) & Fast$_{2}$ (↑) & \# Tokens \\
\midrule

MKEvolve & \textbf{0.99} & 0.77 & 0.49 & \textbf{0.09} & $1.7 \times 10^6$ \\

MKEvolve (Single Split) & 0.98 & \textbf{0.80} & \textbf{0.53} & 0.07 & $1.7 \times 10^6$ \\

\bottomrule
\end{tabular}
\caption{Claude Opus 4.5 Metrics on 100 KernelBench level 2 problems}
\end{subtable}

\vspace{0.5em}

\begin{subtable}{\textwidth}
\centering
\begin{tabular}{lcccccc}
\toprule
\textsc{Method} & Correct (↑) & Fast$_{0.5}$ (↑) & Fast$_{1}$ (↑) & Fast$_{2}$ (↑) & \# Tokens \\
\midrule

MKEvolve & \textbf{0.94} & \textbf{0.74} & 0.34 & \textbf{0.08} & $2.1 \times 10^6$ \\

MKEvolve (Single Split) & 0.90 & 0.62 & \textbf{0.36} & \textbf{0.08} & $2.1 \times 10^6$ \\

\bottomrule
\end{tabular}
\caption{Claude Opus 4.5 Metrics on 50 KernelBench level 3 problems}
\end{subtable}

\caption{Comparison of MKEvolve that iteratively refines the subproblem structure against MKEvolve (Single Split) that decomposes the main problem into subproblems once.}

\end{table}

\section{LLM Inference Kernel Results}
\label{appendix:llmkernels}

Though MKEvolve is primarily intended to advance code generation for large kernels, we also test whether it can produce correct and performant solutions for kernels common in LLM inference.
To this end, we evaluate MKEvolve on three FlashInfer-Bench tasks \citep{xing2026flashinfer}: an FP16 ragged-prefill grouped-query attention task, an FP16 paged-prefill multihead latent attention task, and an FP8 MoE task.

To make the tasks compatible with MKEvolve's LLMDecompose agent, we manually convert them to the KernelBench format in two steps.
First, we wrap FlashInfer-Bench's run function in the forward method of a KernelBench torch module.
Second, we implement KernelBench's \texttt{get\_init\_inputs} function, which returns the constants needed to initialize the module, and its \texttt{get\_inputs} function, which randomly samples fixed-shape input tensors for execution.
For the latter, we take a single input configuration per task from the FlashInfer-Bench dataset and sampled random inputs of those shapes.

Consistent with the FlashInfer-Bench evaluation protocol, we use \texttt{atol=rtol=0.01} for the two attention tasks and \texttt{atol=0.1}, \texttt{rtol=0.2} with an 85\% pass-rate criterion for the MoE task.
We measure all speedups against the \texttt{torch.compile} runtime of each task's PyTorch definition, and compared MKEvolve to two baselines: the FlashInfer-Bench repository's GPT-5 agent, which we re-ran on our A100 machines, and FlashInfer's hand-written CUDA kernels where available.
Given the small number of tasks, we ran the MKEvolve and the GPT-5 agent three times per problem and report the best speedup.

\begin{table}[t]
\centering
\caption{Speedups on FlashInfer-Bench tasks measured against the \texttt{torch.compile} runtime on A100.}
\label{tab:llm-kernel-results}
\small
\begin{tabular}{@{}p{0.65\linewidth}ccc@{}}
\toprule
Task & MKEvolve & GPT-5 Triton & FlashInfer \\
\midrule
\begin{tabular}[t]{@{}l@{}}\texttt{gqa\_ragged\_prefill\_causal\_h32\_kv4\_d128}\end{tabular}
    & $10.7\times$ & $6.0\times$ & $80.0\times$ \\
\begin{tabular}[t]{@{}l@{}}\texttt{mla\_paged\_prefill\_causal\_h16\_ckv512\_kpe64\_ps1}\end{tabular}
    & $44.3\times$ & $23.7\times$ & $238.0\times$ \\
\begin{tabular}[t]{@{}l@{}}\texttt{moe\_fp8\_block\_scale\_ds\_routing\_topk8\_ng8\_kg4\_e32\_h7168\_i2048}\end{tabular}
    & $16.1\times$ & $13.9\times$ & N/A \\
\bottomrule
\end{tabular}
\end{table}

\Cref{tab:llm-kernel-results} reports the results.
Overall, MKEvolve generates Triton kernels that substantially outperform both the \texttt{torch.compile} baseline and the GPT-5 FlashInfer-Bench agent.
The A100 GPU used for these experiments does not natively support FP8, so we found that MKEvolve and GPT-5 solutions dequantize FP8 inputs to higher precision on the fly, at the tile level, during matrix multiplication.
For the same reason, no FlashInfer kernel is available for the FP8 MoE task.
Nevertheless, MKEvolve remains slower than FlashInfer's expert-written, highly optimized CUDA kernels where those are available.
Our primary goal, however, is to study agents that generate Triton code relative to compilers and other inference-time scaling methods, rather than to compete directly with hand-written CUDA: some low-level optimizations are difficult to express in Triton but possible in CUDA.

\section{Example Solution for ConvolutionalVisionTransformer}
\label{appendix:cvtcode}

This section contains a sample MKEvolve output for the KernelBench L3 benchmark's ConvolutionalVisionTransformer problem.
It first lists \textit{decomposed\_solution.py}, which is the end-to-end kernel implementation of the ConvolutionalVisionTransformer module.
Then, it lists all the subkernels in the \textit{subkernel\_implementations} folder imported by \textit{decomposed\_solution.py}.
We highlight that this modular structure allows easy replacement of subkernel implementations in the end-to-end kernel by placing different self-contained subkernel implementations in the \textit{subkernel\_implementations} folder.

\subsection{Top-Level Module}

\begin{nolinenumbers}
\begin{lstlisting}[title={decomposed\_solution.py}]
import sys
sys.path.insert(0, 'outputs_e2e/colm-demo/cvt/c4o-split-prop-fuse-i5f-s0/32_ConvolutionalVisionTransformer/orchestrator/iteration_4')
import torch.nn as nn
from subkernel_implementations.conv2d_patch_embed import ModelNew as Conv2dPatchEmbed
from subkernel_implementations.linear_proj import ModelNew as LinearProj
from subkernel_implementations.transformer_block import ModelNew as TransformerBlock
from subkernel_implementations.cls_token_concat import ModelNew as ClsTokenConcat
from subkernel_implementations.linear_classifier import ModelNew as LinearClassifier

class ModelNew(nn.Module):
    def __init__(self, num_classes, embed_dim=512, num_heads=8, num_layers=6,
                 mlp_ratio=4.0, patch_size=4, in_channels=3, image_size=32):
        super(ModelNew, self).__init__()
        
        self.patch_size = patch_size
        self.image_size = image_size
        self.embed_dim = embed_dim
        
        num_patches = (image_size // patch_size) ** 2
        
        # Match original initialization order exactly
        self.conv1 = Conv2dPatchEmbed(in_channels, embed_dim, patch_size)
        self.linear_proj = LinearProj(embed_dim * num_patches, embed_dim)
        
        self.transformer_layers = []
        for i in range(num_layers):
            # Fused transformer block (self-attention + feedforward)
            # This creates modules in same order: self_attn, then linear1, linear2
            transformer = TransformerBlock(
                d_model=embed_dim,
                nhead=num_heads,
                dim_feedforward=int(embed_dim * mlp_ratio)
            )
            setattr(self, f'transformer_{i}', transformer)
            self.transformer_layers.append(transformer)
        
        self.cls_token_concat = ClsTokenConcat(embed_dim)
        self.fc_out = LinearClassifier(embed_dim, num_classes)
    
    def forward(self, x):
        B = x.size(0)
        x = self.conv1(x)
        x = self.linear_proj(x)
        x = self.cls_token_concat(x, B)
        
        for transformer in self.transformer_layers:
            x = transformer(x)
        
        return self.fc_out(x)
\end{lstlisting}
\end{nolinenumbers}

\subsection{Subkernels}

\begin{nolinenumbers}

\lstinputlisting[title={subkernel\_implementations/conv2d\_patch\_embed.py}]{code_files/ConvolutionalVisionTransformer/subkernel\_implementations/conv2d_patch_embed.py}

\begin{lstlisting}[title={subkernel\_implementations/linear\_proj.py}]
import torch
import torch.nn as nn
import triton
import triton.language as tl
from torch.nn import init


@triton.autotune(
    configs=[
        # Configs optimized for small M (batch=10), large K (8192), medium N (128)
        triton.Config({'BLOCK_SIZE_M': 16, 'BLOCK_SIZE_N': 32, 'BLOCK_SIZE_K': 128, 'GROUP_SIZE_M': 8}, num_stages=4, num_warps=4),
        triton.Config({'BLOCK_SIZE_M': 16, 'BLOCK_SIZE_N': 64, 'BLOCK_SIZE_K': 128, 'GROUP_SIZE_M': 8}, num_stages=4, num_warps=4),
        triton.Config({'BLOCK_SIZE_M': 16, 'BLOCK_SIZE_N': 128, 'BLOCK_SIZE_K': 64, 'GROUP_SIZE_M': 8}, num_stages=4, num_warps=4),
        triton.Config({'BLOCK_SIZE_M': 16, 'BLOCK_SIZE_N': 128, 'BLOCK_SIZE_K': 128, 'GROUP_SIZE_M': 8}, num_stages=3, num_warps=4),
        triton.Config({'BLOCK_SIZE_M': 32, 'BLOCK_SIZE_N': 32, 'BLOCK_SIZE_K': 128, 'GROUP_SIZE_M': 8}, num_stages=4, num_warps=4),
        triton.Config({'BLOCK_SIZE_M': 32, 'BLOCK_SIZE_N': 64, 'BLOCK_SIZE_K': 64, 'GROUP_SIZE_M': 8}, num_stages=4, num_warps=4),
        triton.Config({'BLOCK_SIZE_M': 32, 'BLOCK_SIZE_N': 128, 'BLOCK_SIZE_K': 64, 'GROUP_SIZE_M': 8}, num_stages=4, num_warps=4),
    ],
    key=['M', 'N', 'K'],
)
@triton.jit
def linear_kernel(
    x_ptr, wt_ptr, b_ptr, y_ptr,
    M, N, K,
    stride_xm, stride_xk,
    stride_wtk, stride_wtn,
    stride_ym, stride_yn,
    BLOCK_SIZE_M: tl.constexpr, BLOCK_SIZE_N: tl.constexpr, BLOCK_SIZE_K: tl.constexpr,
    GROUP_SIZE_M: tl.constexpr,
):
    """
    Fused linear layer kernel: y = x @ W_T + b
    where W_T is the pre-transposed weight (K, N)
    x: (M, K), W_T: (K, N), b: (N,), y: (M, N)
    """
    pid = tl.program_id(axis=0)
    
    num_pid_m = tl.cdiv(M, BLOCK_SIZE_M)
    num_pid_n = tl.cdiv(N, BLOCK_SIZE_N)
    
    num_pid_in_group = GROUP_SIZE_M * num_pid_n
    group_id = pid // num_pid_in_group
    first_pid_m = group_id * GROUP_SIZE_M
    group_size_m = min(num_pid_m - first_pid_m, GROUP_SIZE_M)
    pid_m = first_pid_m + (pid % group_size_m)
    pid_n = (pid % num_pid_in_group) // group_size_m
    
    offs_m = pid_m * BLOCK_SIZE_M + tl.arange(0, BLOCK_SIZE_M)
    offs_n = pid_n * BLOCK_SIZE_N + tl.arange(0, BLOCK_SIZE_N)
    offs_k = tl.arange(0, BLOCK_SIZE_K)
    
    x_ptrs = x_ptr + offs_m[:, None] * stride_xm + offs_k[None, :] * stride_xk
    wt_ptrs = wt_ptr + offs_k[:, None] * stride_wtk + offs_n[None, :] * stride_wtn
    
    acc = tl.zeros((BLOCK_SIZE_M, BLOCK_SIZE_N), dtype=tl.float32)
    
    for k in range(0, tl.cdiv(K, BLOCK_SIZE_K)):
        k_remaining = K - k * BLOCK_SIZE_K
        k_mask = offs_k < k_remaining
        
        x_block = tl.load(x_ptrs, mask=(offs_m[:, None] < M) & k_mask[None, :], other=0.0)
        wt_block = tl.load(wt_ptrs, mask=k_mask[:, None] & (offs_n[None, :] < N), other=0.0)
        
        acc = tl.dot(x_block, wt_block, acc, allow_tf32=False)
        
        x_ptrs += BLOCK_SIZE_K * stride_xk
        wt_ptrs += BLOCK_SIZE_K * stride_wtk
    
    # Fused bias addition
    bias = tl.load(b_ptr + offs_n, mask=offs_n < N, other=0.0)
    acc = acc + bias[None, :]
    
    # Store output
    offs_m_out = pid_m * BLOCK_SIZE_M + tl.arange(0, BLOCK_SIZE_M)
    offs_n_out = pid_n * BLOCK_SIZE_N + tl.arange(0, BLOCK_SIZE_N)
    y_ptrs = y_ptr + offs_m_out[:, None] * stride_ym + offs_n_out[None, :] * stride_yn
    mask = (offs_m_out[:, None] < M) & (offs_n_out[None, :] < N)
    tl.store(y_ptrs, acc, mask=mask)


class ModelNew(nn.Module):
    def __init__(self, in_features, out_features):
        super().__init__()
        self.in_features = in_features
        self.out_features = out_features
        
        weight = torch.empty((out_features, in_features))
        init.kaiming_uniform_(weight, a=5**0.5)
        
        bias = torch.empty(out_features)
        fan_in, _ = init._calculate_fan_in_and_fan_out(weight)
        bound = 1 / fan_in**0.5 if fan_in > 0 else 0
        init.uniform_(bias, -bound, bound)
        
        # Store weight transposed for efficient memory access: (K, N) instead of (N, K)
        self.weight_t = nn.Parameter(weight.t().contiguous())
        self.bias = nn.Parameter(bias)
    
    def forward(self, x):
        M = x.shape[0]
        K = self.in_features
        N = self.out_features
        
        x = x.contiguous()
        y = torch.empty((M, N), device=x.device, dtype=x.dtype)
        
        grid = lambda META: (triton.cdiv(M, META['BLOCK_SIZE_M']) * triton.cdiv(N, META['BLOCK_SIZE_N']),)
        
        linear_kernel[grid](
            x, self.weight_t, self.bias, y,
            M, N, K,
            x.stride(0), x.stride(1),
            self.weight_t.stride(0), self.weight_t.stride(1),
            y.stride(0), y.stride(1),
        )
        
        return y
\end{lstlisting}

\begin{lstlisting}[title={subkernel\_implementations/cls\_token\_concat.py}]
import triton
import triton.language as tl
import torch
import torch.nn as nn


@triton.autotune(
    configs=[
        triton.Config({'BLOCK_SIZE': 1024}, num_warps=4),
        triton.Config({'BLOCK_SIZE': 2048}, num_warps=8),
        triton.Config({'BLOCK_SIZE': 512}, num_warps=2),
        triton.Config({'BLOCK_SIZE': 256}, num_warps=2),
    ],
    key=['total_elements'],
)
@triton.jit
def cls_concat_kernel_1d(
    x_ptr,           # Input tensor (B, embed_dim)
    cls_ptr,         # CLS token (embed_dim,) flattened
    out_ptr,         # Output tensor (B, 2, embed_dim)
    batch_size,      # B
    embed_dim,       # embed_dim
    total_elements,  # B * 2 * embed_dim
    BLOCK_SIZE: tl.constexpr,
):
    """
    1D kernel that writes output directly.
    Output layout: [batch, seq, embed] where seq=2
    Linear index i maps to (b, s, e) where:
      - e = i % embed_dim
      - s = (i // embed_dim) % 2
      - b = i // (2 * embed_dim)
    """
    pid = tl.program_id(0)
    
    # Calculate linear offsets for this block
    offs = pid * BLOCK_SIZE + tl.arange(0, BLOCK_SIZE)
    mask = offs < total_elements
    
    # Decompose linear index to (batch, seq, embed)
    embed_dim_val = embed_dim
    seq_stride = embed_dim_val
    batch_stride = 2 * embed_dim_val
    
    e = offs % embed_dim_val
    s = (offs // seq_stride) % 2
    b = offs // batch_stride
    
    # For s=0: load from cls_token[e]
    # For s=1: load from x[b, e] = x[b * embed_dim + e]
    
    # Load cls values (only need embed dimension)
    cls_vals = tl.load(cls_ptr + e, mask=mask)
    
    # Load x values
    x_offs = b * embed_dim_val + e
    x_vals = tl.load(x_ptr + x_offs, mask=mask & (b < batch_size))
    
    # Select based on sequence position
    out_vals = tl.where(s == 0, cls_vals, x_vals)
    
    # Store to output
    tl.store(out_ptr + offs, out_vals, mask=mask)


class ModelNew(nn.Module):
    """
    Optimized Triton implementation of cls token concatenation.
    Uses a 1D kernel with autotuning for better performance.
    """
    
    def __init__(self, embed_dim):
        super().__init__()
        self.cls_token = nn.Parameter(torch.zeros(1, 1, embed_dim))
    
    def forward(self, x, batch_size):
        B = x.shape[0]
        embed_dim = x.shape[1]
        
        # Allocate output tensor - already contiguous
        out = torch.empty((B, 2, embed_dim), device=x.device, dtype=x.dtype)
        
        # Flatten cls_token for simple 1D access
        cls_flat = self.cls_token.view(-1)
        
        # Total elements in output
        total_elements = B * 2 * embed_dim
        
        # Grid: number of blocks needed
        def grid(META):
            return (triton.cdiv(total_elements, META['BLOCK_SIZE']),)
        
        # Launch kernel
        cls_concat_kernel_1d[grid](
            x,
            cls_flat,
            out,
            B,
            embed_dim,
            total_elements,
        )
        
        return out
\end{lstlisting}

\begin{lstlisting}[title={subkernel\_implementations/transformer\_block.py}]
import triton
import triton.language as tl
import torch
import torch.nn as nn
from torch.nn import init
import math

@triton.autotune(
    configs=[
        triton.Config({'BLOCK_BS': 32, 'BLOCK_D_IN': 64, 'BLOCK_D_OUT': 64}, num_warps=4),
        triton.Config({'BLOCK_BS': 32, 'BLOCK_D_IN': 32, 'BLOCK_D_OUT': 128}, num_warps=4),
        triton.Config({'BLOCK_BS': 16, 'BLOCK_D_IN': 64, 'BLOCK_D_OUT': 128}, num_warps=4),
        triton.Config({'BLOCK_BS': 64, 'BLOCK_D_IN': 32, 'BLOCK_D_OUT': 64}, num_warps=4),
        triton.Config({'BLOCK_BS': 16, 'BLOCK_D_IN': 32, 'BLOCK_D_OUT': 64}, num_warps=2),
    ],
    key=['batch_seq', 'd_model'],
)
@triton.jit
def qkv_projection_kernel(
    x_ptr, w_ptr, b_ptr, out_ptr,
    batch_seq, d_model, d_model_3,
    stride_x_bs, stride_x_d, stride_w_out, stride_w_in, stride_out_bs, stride_out_d,
    BLOCK_BS: tl.constexpr, BLOCK_D_IN: tl.constexpr, BLOCK_D_OUT: tl.constexpr,
):
    pid_bs = tl.program_id(0)
    pid_d = tl.program_id(1)
    offs_bs = pid_bs * BLOCK_BS + tl.arange(0, BLOCK_BS)
    offs_d_out = pid_d * BLOCK_D_OUT + tl.arange(0, BLOCK_D_OUT)
    acc = tl.zeros((BLOCK_BS, BLOCK_D_OUT), dtype=tl.float32)
    for k in range(0, d_model, BLOCK_D_IN):
        offs_k = k + tl.arange(0, BLOCK_D_IN)
        x = tl.load(x_ptr + offs_bs[:, None] * stride_x_bs + offs_k[None, :],
                    mask=(offs_bs[:, None] < batch_seq) & (offs_k[None, :] < d_model), other=0.0)
        w = tl.load(w_ptr + offs_d_out[:, None] * stride_w_out + offs_k[None, :],
                    mask=(offs_d_out[:, None] < d_model_3) & (offs_k[None, :] < d_model), other=0.0)
        acc += tl.dot(x, tl.trans(w), allow_tf32=False)
    bias = tl.load(b_ptr + offs_d_out, mask=offs_d_out < d_model_3, other=0.0)
    acc = acc + bias[None, :]
    tl.store(out_ptr + offs_bs[:, None] * stride_out_bs + offs_d_out[None, :],
             acc.to(out_ptr.dtype.element_ty), mask=(offs_bs[:, None] < batch_seq) & (offs_d_out[None, :] < d_model_3))

@triton.jit
def attention_kernel(
    qkv_ptr, out_ptr, batch, seq_len, nhead, head_dim, d_model, scale,
    stride_qkv_bs, stride_o_bs,
    BLOCK_S: tl.constexpr, BLOCK_D: tl.constexpr,
):
    pid_bh = tl.program_id(0)
    pid_s = tl.program_id(1)
    batch_idx = pid_bh // nhead
    head_idx = pid_bh % nhead
    offs_s = pid_s * BLOCK_S + tl.arange(0, BLOCK_S)
    offs_d = tl.arange(0, BLOCK_D)
    
    base_bs = batch_idx * seq_len
    q_off = head_idx * head_dim
    k_off = d_model + head_idx * head_dim
    v_off = 2 * d_model + head_idx * head_dim
    
    q = tl.load(qkv_ptr + (base_bs + offs_s[:, None]) * stride_qkv_bs + q_off + offs_d[None, :],
                mask=(offs_s[:, None] < seq_len) & (offs_d[None, :] < head_dim), other=0.0).to(tl.float32)
    
    row_max = tl.full((BLOCK_S,), float('-inf'), dtype=tl.float32)
    row_sum = tl.zeros((BLOCK_S,), dtype=tl.float32)
    acc = tl.zeros((BLOCK_S, BLOCK_D), dtype=tl.float32)
    
    for j in range(0, seq_len, BLOCK_S):
        offs_j = j + tl.arange(0, BLOCK_S)
        k = tl.load(qkv_ptr + (base_bs + offs_j[:, None]) * stride_qkv_bs + k_off + offs_d[None, :],
                    mask=(offs_j[:, None] < seq_len) & (offs_d[None, :] < head_dim), other=0.0).to(tl.float32)
        qk = tl.dot(q, tl.trans(k), allow_tf32=False) * scale
        qk = tl.where((offs_s[:, None] < seq_len) & (offs_j[None, :] < seq_len), qk, float('-inf'))
        new_max = tl.maximum(row_max, tl.max(qk, axis=1))
        exp_old = tl.exp(row_max - new_max)
        exp_qk = tl.exp(qk - new_max[:, None])
        new_sum = row_sum * exp_old + tl.sum(exp_qk, axis=1)
        acc = acc * exp_old[:, None]
        v = tl.load(qkv_ptr + (base_bs + offs_j[:, None]) * stride_qkv_bs + v_off + offs_d[None, :],
                    mask=(offs_j[:, None] < seq_len) & (offs_d[None, :] < head_dim), other=0.0).to(tl.float32)
        acc += tl.dot(exp_qk.to(tl.float32), v, allow_tf32=False)
        row_max = new_max
        row_sum = new_sum
    
    acc = acc / row_sum[:, None]
    tl.store(out_ptr + (base_bs + offs_s[:, None]) * stride_o_bs + q_off + offs_d[None, :],
             acc.to(out_ptr.dtype.element_ty), mask=(offs_s[:, None] < seq_len) & (offs_d[None, :] < head_dim))

@triton.jit
def proj_residual_ln_kernel(
    attn_ptr, x_ptr, w_ptr, b_ptr, ln_w_ptr, ln_b_ptr, out_ptr,
    batch_seq, d_model, eps,
    BLOCK_D: tl.constexpr,
):
    pid = tl.program_id(0)
    offs_d = tl.arange(0, BLOCK_D)
    d_mask = offs_d < d_model
    proj_acc = tl.zeros((BLOCK_D,), dtype=tl.float32)
    for k in range(0, d_model, BLOCK_D):
        offs_k = k + tl.arange(0, BLOCK_D)
        k_mask = offs_k < d_model
        a = tl.load(attn_ptr + pid * d_model + offs_k, mask=k_mask, other=0.0).to(tl.float32)
        w = tl.load(w_ptr + offs_d[:, None] * d_model + offs_k[None, :], mask=d_mask[:, None] & k_mask[None, :], other=0.0).to(tl.float32)
        proj_acc += tl.sum(w * a[None, :], axis=1)
    proj_acc = proj_acc + tl.load(b_ptr + offs_d, mask=d_mask, other=0.0).to(tl.float32)
    x = tl.load(x_ptr + pid * d_model + offs_d, mask=d_mask, other=0.0).to(tl.float32)
    residual = proj_acc + x
    mean = tl.sum(residual, axis=0) / d_model
    var = tl.sum((residual - mean) * (residual - mean), axis=0) / d_model
    rstd = 1.0 / tl.sqrt(var + eps)
    ln_w = tl.load(ln_w_ptr + offs_d, mask=d_mask, other=1.0).to(tl.float32)
    ln_b = tl.load(ln_b_ptr + offs_d, mask=d_mask, other=0.0).to(tl.float32)
    tl.store(out_ptr + pid * d_model + offs_d, ((residual - mean) * rstd * ln_w + ln_b).to(out_ptr.dtype.element_ty), mask=d_mask)

@triton.autotune(
    configs=[
        triton.Config({'BLOCK_M': 32, 'BLOCK_N': 128, 'BLOCK_K': 32}, num_warps=4),
        triton.Config({'BLOCK_M': 16, 'BLOCK_N': 256, 'BLOCK_K': 32}, num_warps=4),
        triton.Config({'BLOCK_M': 32, 'BLOCK_N': 256, 'BLOCK_K': 32}, num_warps=8),
        triton.Config({'BLOCK_M': 16, 'BLOCK_N': 128, 'BLOCK_K': 64}, num_warps=4),
    ],
    key=['M', 'N'],
)
@triton.jit
def linear_relu_kernel(
    x_ptr, w_ptr, b_ptr, out_ptr, M, K, N,
    stride_xm, stride_wn, stride_outm,
    BLOCK_M: tl.constexpr, BLOCK_N: tl.constexpr, BLOCK_K: tl.constexpr,
):
    pid_m, pid_n = tl.program_id(0), tl.program_id(1)
    offs_m = pid_m * BLOCK_M + tl.arange(0, BLOCK_M)
    offs_n = pid_n * BLOCK_N + tl.arange(0, BLOCK_N)
    acc = tl.zeros((BLOCK_M, BLOCK_N), dtype=tl.float32)
    for k in range(0, K, BLOCK_K):
        offs_k = k + tl.arange(0, BLOCK_K)
        x_block = tl.load(x_ptr + offs_m[:, None] * stride_xm + offs_k[None, :], mask=(offs_m[:, None] < M) & (offs_k[None, :] < K), other=0.0)
        w_block = tl.load(w_ptr + offs_n[None, :] * stride_wn + offs_k[:, None], mask=(offs_n[None, :] < N) & (offs_k[:, None] < K), other=0.0)
        acc += tl.dot(x_block, w_block, allow_tf32=False)
    acc = tl.maximum(acc + tl.load(b_ptr + offs_n, mask=offs_n < N, other=0.0)[None, :], 0.0)
    tl.store(out_ptr + offs_m[:, None] * stride_outm + offs_n[None, :], acc, mask=(offs_m[:, None] < M) & (offs_n[None, :] < N))

@triton.jit
def linear_res_ln_kernel(
    h_ptr, w_ptr, b_ptr, x_ptr, g_ptr, bt_ptr, out_ptr,
    M, K, N, eps,
    BLOCK_N: tl.constexpr, BLOCK_K: tl.constexpr,
):
    pid = tl.program_id(0)
    offs_n = tl.arange(0, BLOCK_N)
    n_mask = offs_n < N
    ff = tl.zeros((BLOCK_N,), dtype=tl.float32)
    for k in range(0, K, BLOCK_K):
        offs_k = k + tl.arange(0, BLOCK_K)
        k_mask = offs_k < K
        h = tl.load(h_ptr + pid * K + offs_k, mask=k_mask, other=0.0)
        w = tl.load(w_ptr + offs_n[:, None] * K + offs_k[None, :], mask=n_mask[:, None] & k_mask[None, :], other=0.0)
        ff += tl.sum(w * h[None, :], axis=1)
    res = tl.load(x_ptr + pid * N + offs_n, mask=n_mask, other=0.0) + ff + tl.load(b_ptr + offs_n, mask=n_mask, other=0.0)
    mean = tl.sum(tl.where(n_mask, res, 0.0), axis=0) / N
    diff = tl.where(n_mask, res - mean, 0.0)
    rstd = 1.0 / tl.sqrt(tl.sum(diff * diff, axis=0) / N + eps)
    out = (res - mean) * rstd * tl.load(g_ptr + offs_n, mask=n_mask, other=1.0) + tl.load(bt_ptr + offs_n, mask=n_mask, other=0.0)
    tl.store(out_ptr + pid * N + offs_n, out, mask=n_mask)

class ModelNew(nn.Module):
    def __init__(self, d_model, nhead, dim_feedforward):
        super().__init__()
        self.d_model, self.nhead, self.head_dim, self.dim_feedforward = d_model, nhead, d_model // nhead, dim_feedforward
        self.out_proj_weight, self.out_proj_bias = nn.Parameter(torch.empty(d_model, d_model)), nn.Parameter(torch.empty(d_model))
        init.kaiming_uniform_(self.out_proj_weight, a=math.sqrt(5))
        fan_in, _ = init._calculate_fan_in_and_fan_out(self.out_proj_weight)
        init.uniform_(self.out_proj_bias, -1/math.sqrt(fan_in), 1/math.sqrt(fan_in))
        self.in_proj_weight, self.in_proj_bias = nn.Parameter(torch.empty(3*d_model, d_model)), nn.Parameter(torch.empty(3*d_model))
        init.xavier_uniform_(self.in_proj_weight); init.constant_(self.in_proj_bias, 0.0); init.constant_(self.out_proj_bias, 0.0)
        self.ln1_weight, self.ln1_bias = nn.Parameter(torch.ones(d_model)), nn.Parameter(torch.zeros(d_model))
        self.weight1, self.bias1 = nn.Parameter(torch.empty(dim_feedforward, d_model)), nn.Parameter(torch.empty(dim_feedforward))
        init.kaiming_uniform_(self.weight1, a=math.sqrt(5)); fan_in1, _ = init._calculate_fan_in_and_fan_out(self.weight1); init.uniform_(self.bias1, -1/math.sqrt(fan_in1), 1/math.sqrt(fan_in1))
        self.weight2, self.bias2 = nn.Parameter(torch.empty(d_model, dim_feedforward)), nn.Parameter(torch.empty(d_model))
        init.kaiming_uniform_(self.weight2, a=math.sqrt(5)); fan_in2, _ = init._calculate_fan_in_and_fan_out(self.weight2); init.uniform_(self.bias2, -1/math.sqrt(fan_in2), 1/math.sqrt(fan_in2))
        self.ln2_weight, self.ln2_bias = nn.Parameter(torch.ones(d_model)), nn.Parameter(torch.zeros(d_model))

    def forward(self, x):
        B, S, D = x.shape; BS = B * S
        x_flat = x.view(BS, D)
        qkv = torch.empty(BS, 3*D, device=x.device, dtype=x.dtype)
        grid_qkv = lambda META: (triton.cdiv(BS, META['BLOCK_BS']), triton.cdiv(3*D, META['BLOCK_D_OUT']))
        qkv_projection_kernel[grid_qkv](x_flat, self.in_proj_weight, self.in_proj_bias, qkv, BS, D, 3*D, D, 1, D, 1, 3*D, 1)
        attn = torch.empty(BS, D, device=x.device, dtype=x.dtype)
        BLOCK_S, BLOCK_D = max(16, triton.next_power_of_2(S)), max(16, triton.next_power_of_2(self.head_dim))
        attention_kernel[(B*self.nhead, triton.cdiv(S, BLOCK_S))](qkv, attn, B, S, self.nhead, self.head_dim, D, 1.0/math.sqrt(self.head_dim), 3*D, D, BLOCK_S, BLOCK_D)
        attn_n = torch.empty(BS, D, device=x.device, dtype=x.dtype)
        proj_residual_ln_kernel[(BS,)](attn, x_flat, self.out_proj_weight, self.out_proj_bias, self.ln1_weight, self.ln1_bias, attn_n, BS, D, 1e-5, max(128, triton.next_power_of_2(D)))
        hidden = torch.empty(BS, self.dim_feedforward, device=x.device, dtype=x.dtype)
        grid1 = lambda META: (triton.cdiv(BS, META['BLOCK_M']), triton.cdiv(self.dim_feedforward, META['BLOCK_N']))
        linear_relu_kernel[grid1](attn_n, self.weight1, self.bias1, hidden, BS, D, self.dim_feedforward, D, D, self.dim_feedforward)
        out = torch.empty(BS, D, device=x.device, dtype=x.dtype)
        linear_res_ln_kernel[(BS,)](hidden, self.weight2, self.bias2, attn_n, self.ln2_weight, self.ln2_bias, out, BS, self.dim_feedforward, D, 1e-5, triton.next_power_of_2(D), 128)
        return out.view(B, S, D)
\end{lstlisting}

\begin{lstlisting}[title={subkernel\_implementations/classification\_head.py}]
import torch
import torch.nn as nn
import triton
import triton.language as tl
from torch.nn import init

@triton.autotune(
    configs=[
        triton.Config({'BLOCK_SIZE_M': 16, 'BLOCK_SIZE_N': 64, 'BLOCK_SIZE_K': 32}, num_warps=4),
        triton.Config({'BLOCK_SIZE_M': 16, 'BLOCK_SIZE_N': 128, 'BLOCK_SIZE_K': 32}, num_warps=4),
        triton.Config({'BLOCK_SIZE_M': 32, 'BLOCK_SIZE_N': 64, 'BLOCK_SIZE_K': 32}, num_warps=4),
        triton.Config({'BLOCK_SIZE_M': 32, 'BLOCK_SIZE_N': 128, 'BLOCK_SIZE_K': 64}, num_warps=4),
    ],
    key=['M', 'N', 'K'],
)
@triton.jit
def fused_cls_linear_kernel(
    x_ptr,          # Input tensor (B, 2, embed_dim)
    weight_ptr,     # Weight matrix (num_classes, embed_dim)
    bias_ptr,       # Bias vector (num_classes,)
    output_ptr,     # Output tensor (B, num_classes)
    M,              # Batch size
    N,              # num_classes
    K,              # embed_dim
    stride_x_b,     # Stride for batch dimension in x
    stride_x_seq,   # Stride for sequence dimension in x
    stride_x_k,     # Stride for embed dimension in x
    stride_w_n,     # Stride for output dimension in weight
    stride_w_k,     # Stride for input dimension in weight
    stride_out_m,   # Stride for batch dimension in output
    stride_out_n,   # Stride for class dimension in output
    BLOCK_SIZE_M: tl.constexpr,
    BLOCK_SIZE_N: tl.constexpr,
    BLOCK_SIZE_K: tl.constexpr,
):
    """
    Fused kernel that extracts CLS token (position 0) and applies linear transformation.
    Computes: output[b, n] = sum_k(x[b, 0, k] * weight[n, k]) + bias[n]
    """
    pid_m = tl.program_id(0)  # Batch block
    pid_n = tl.program_id(1)  # Output class block
    
    # Offsets for this block
    offs_m = pid_m * BLOCK_SIZE_M + tl.arange(0, BLOCK_SIZE_M)
    offs_n = pid_n * BLOCK_SIZE_N + tl.arange(0, BLOCK_SIZE_N)
    offs_k = tl.arange(0, BLOCK_SIZE_K)
    
    # Masks for boundary conditions
    mask_m = offs_m < M
    mask_n = offs_n < N
    
    # Initialize accumulator
    acc = tl.zeros((BLOCK_SIZE_M, BLOCK_SIZE_N), dtype=tl.float32)
    
    # Loop over K dimension
    for k_start in range(0, K, BLOCK_SIZE_K):
        k_offs = k_start + offs_k
        mask_k = k_offs < K
        
        # Load x[:, 0, k] - CLS token for each batch element
        # x_ptr points to x[b, 0, k] = x_ptr + b * stride_x_b + 0 * stride_x_seq + k * stride_x_k
        x_ptrs = x_ptr + offs_m[:, None] * stride_x_b + k_offs[None, :] * stride_x_k
        x_block = tl.load(x_ptrs, mask=mask_m[:, None] & mask_k[None, :], other=0.0)
        
        # Load weight[n, k]
        w_ptrs = weight_ptr + offs_n[:, None] * stride_w_n + k_offs[None, :] * stride_w_k
        w_block = tl.load(w_ptrs, mask=mask_n[:, None] & mask_k[None, :], other=0.0)
        
        # Accumulate: x_block is (M, K), w_block is (N, K)
        # We need x @ w.T, so x_block @ w_block.T
        acc += tl.dot(x_block, tl.trans(w_block), allow_tf32=False)
    
    # Load and add bias
    bias = tl.load(bias_ptr + offs_n, mask=mask_n, other=0.0)
    acc = acc + bias[None, :]
    
    # Store output
    out_ptrs = output_ptr + offs_m[:, None] * stride_out_m + offs_n[None, :] * stride_out_n
    tl.store(out_ptrs, acc, mask=mask_m[:, None] & mask_n[None, :])


class ModelNew(nn.Module):
    """
    Triton implementation of CLS token extraction followed by linear layer.
    Fuses: x[:, 0] extraction + matmul + bias addition into a single kernel.
    """
    def __init__(self, in_features, num_classes):
        super().__init__()
        self.in_features = in_features
        self.num_classes = num_classes
        
        # Initialize weight and bias exactly as nn.Linear does
        weight = torch.empty((num_classes, in_features))
        init.kaiming_uniform_(weight, a=5**0.5)
        
        bias = torch.empty(num_classes)
        fan_in, _ = init._calculate_fan_in_and_fan_out(weight)
        bound = 1 / fan_in**0.5 if fan_in > 0 else 0
        init.uniform_(bias, -bound, bound)
        
        self.weight = nn.Parameter(weight)
        self.bias = nn.Parameter(bias)
    
    def forward(self, x):
        """
        Args:
            x: Input tensor of shape (B, 2, embed_dim)
        Returns:
            Output tensor of shape (B, num_classes)
        """
        B, seq_len, embed_dim = x.shape
        M = B
        N = self.num_classes
        K = self.in_features
        
        # Allocate output
        output = torch.empty((M, N), device=x.device, dtype=x.dtype)
        
        # Grid calculation
        grid = lambda META: (
            triton.cdiv(M, META['BLOCK_SIZE_M']),
            triton.cdiv(N, META['BLOCK_SIZE_N']),
        )
        
        # Launch kernel
        fused_cls_linear_kernel[grid](
            x, self.weight, self.bias, output,
            M, N, K,
            x.stride(0), x.stride(1), x.stride(2),
            self.weight.stride(0), self.weight.stride(1),
            output.stride(0), output.stride(1),
        )
        
        return output
\end{lstlisting}


\end{nolinenumbers}

\section{Prompts}

\subsection{Subproblem Decomposition}

\begin{Verbatim}[breaklines=true]
Decompose the PyTorch module under the "PROBLEM TO DECOMPOSE" section into submodules optimized for kernel fusion.

## TASK OBJECTIVE

Create a `ModelNew` nn.Module in `decomposed_problem.py` that is **functionally equivalent** to the `Model` class in the problem description below. The `ModelNew` module should:
- Accept the same inputs as the original `Model`
- Produce numerically identical outputs
- Be composed of submodules from `subkernel_problems/` that can each be independently optimized as fused kernels
- Not use any other torch.nn modules in its code than submodules from `subkernel_problems/`, including modules such as `nn.ModuleList` and `nn.Parameters`
- Do all mathematical computations in the files within `subkernel_problems/` and have no mathematical computations directly in the file `decomposed_problem.py`
- Avoid creating submodules that only perform highly trivial operations such as no-op, identity function, copy function, contiguous operation, flatten operation, packing and unpacking data, etc
- Avoid creating submodules that pass one torch.nn Module as an argument to another torch.nn Module, the only exceptions applying to nn.Sequential, nn.ModuleList, and nn.ModuleDict

Note on Numerical Equivalence: PyTorch uses a global random number generator for weight initialization. When `ModelNew.__init__` instantiates subkernel modules (from `subkernel_problems/`), those subkernel modules internally create `torch.nn` modules. The order in which `ModelNew.__init__` instantiates subkernel modules must ensure the underlying `torch.nn` modules are created in the EXACT SAME ORDER as they appear in the original `Model.__init__`. For example, if the original model has:
```python
self.conv1 = nn.Conv2d(3, 64, 3)   # consumes RNG state first
self.bn1 = nn.BatchNorm2d(64)      # consumes RNG state second
self.conv2 = nn.Conv2d(64, 128, 3) # consumes RNG state third
```
Then `ModelNew.__init__` must instantiate its subkernel modules in an order that causes Conv2d(3,64,3) to be created first, then BatchNorm2d(64), then Conv2d(64,128,3). Misordering causes different random values to be assigned to weights, resulting in numerical mismatch.
This also applies when decomposing a submodule containing a single torch.nn layer that contains multiple torch.nn layers such as nn.MultiheadAttention. For reference, its parameter initialization method is
```python
def _reset_parameters(self):
    if self._qkv_same_embed_dim:
        xavier_uniform_(self.in_proj_weight)
    else:
        xavier_uniform_(self.q_proj_weight)
        xavier_uniform_(self.k_proj_weight)
        xavier_uniform_(self.v_proj_weight)
    if self.in_proj_bias is not None:
        constant_(self.in_proj_bias, 0.0)
        constant_(self.out_proj.bias, 0.0)
    if self.bias_k is not None:
        xavier_normal_(self.bias_k)
    if self.bias_v is not None:
        xavier_normal_(self.bias_v)
```

## VERIFICATION CRITERIA

Your decomposition will be validated against these checks:

1. **Functional Equivalence**: `ModelNew(*get_inputs())` must produce the same output as `Model(*get_inputs())` from the original problem file (numerical equivalence check)
2. **Subkernel Standalone Execution**: Each subkernel in `subkernel_problems/` must execute successfully with its own `get_inputs()` and `get_init_inputs()` (or `get_input_configs()`):
   - `model = Model(*get_init_inputs())` must instantiate without error
   - `output = model(*get_inputs())` must execute without error
3. **Shape Consistency**: The `get_inputs()` in each subkernel file must return tensors with shapes matching what `ModelNew.forward()` will actually pass to that submodule
4. **Init Function Consistency**: The `get_init_inputs()` in each subkernel file must return arguments that `ModelNew` uses to initialize the corresponding `Model` module
5. **Input Function Coverage**: If `ModelNew` nn.Module at `decomposed_problem.py` contain multiple instantiation of subkernel PyTorch modules in `subkernel_problems`, the corresponding subkernel PyTorch module file must have the function `get_input_configs()` that returns a list of dicts with functions that return `get_init_inputs()` and `get_inputs()` shapes present in `ModelNew`'s forward pass
6. **Cheating Detection**: The only PyTorch modules initialized at `decomposed_problem.py` should be modules from `subkernel_problems/`

## FUSION OBJECTIVE

Maximize fusion opportunities by grouping operations that can be efficiently executed in a single kernel. Only split into separate submodules when fusion is blocked by correctness or hardware constraints.

**Splitting Constraints (only split when these apply):**
- Data dependencies crossing incompatible tiling axes (e.g., reduction reuse across incompatible blocks)
- Operations requiring global synchronization (e.g., full tensor reductions before the result is used)
- Branching control flow that cannot be predicated

**When splitting is necessary:**
- Emit the minimal number of submodules
- Add a comment explaining the blocking constraint (e.g., "reduction requires global synchronization")

## FUSION-FRIENDLY GROUPINGS

**Operations that SHOULD be fused together (single submodule):**
- Matmul/GEMM followed by element-wise ops (e.g., matmul → relu, linear → gelu)
- Chains of element-wise operations (relu, sigmoid, tanh, add, mul, scale)
- Reshape/view/permute operations (free at kernel level)
- Dropout, masking, scaling applied to computation results
- Bias addition after matmul

**Common Fusible Patterns:**
- `Linear + Activation`: nn.Linear → ReLU/GELU/SiLU (single kernel)
- `Attention QKV`: Q@K^T → scale → mask → softmax → @V (may need split at softmax due to reduction)
- `MLP Block`: Linear → Activation → Linear (split between the two linears if needed)
- `Residual Add`: computation + skip connection (fuse the add)

## MEMORY & DATAFLOW OPTIMIZATION

- Treat global memory traffic as the primary cost
- Never materialize intermediates unless unavoidable
- Fuse producers → consumers when data reuse exists
- Keep intermediates in registers/SRAM where possible

## DECOMPOSITION STRATEGY

1. Analyze the forward pass as a dataflow graph
2. Identify operations requiring global synchronization (fusion blockers)
3. Group all other operations between blockers into single submodules
4. Each submodule = one kernel opportunity
5. Ensure exact output semantics, dtype behavior, and broadcasting rules are preserved

## SUBKERNEL INPUT FUNCTION REQUIREMENTS

**CRITICAL DECISION RULE:**
For each subkernel file, count how many times `ModelNew.__init__` instantiates that subkernel's Model class:
- **1 instantiation** → define `get_inputs()` and `get_init_inputs()`
- **N instantiations (N > 1)** → define `get_input_configs()` returning exactly N dicts

**Function Definitions:**

When Model is instantiated ONCE:
- **get_init_inputs()**: Returns tuple of arguments for `Model(*get_init_inputs())`
- **get_inputs()**: Returns tuple of input tensors for `model(*get_inputs())`

When Model is instantiated MULTIPLE TIMES:
- **get_input_configs()**: Returns a list of N dicts (one per instantiation), each containing:
  - `"get_init_inputs"`: function returning tuple for that instance's initialization
  - `"get_inputs"`: function returning tuple for that instance's forward pass

**Shape Requirements:**
- `get_init_inputs()` must return arguments matching what `ModelNew.__init__` passes when instantiating
- `get_inputs()` must return tensors with shapes matching what `ModelNew.forward()` passes to that submodule

This ensures each subkernel can be independently tested and optimized with realistic input shapes.

## SUBMODULE REUSE REQUIREMENT

**Do NOT create duplicate subkernel files for the same operation type, even if they have different shapes.**
Each subkernel file should define a **shape-parameterized** `Model` class that can be instantiated multiple times with different dimension arguments. A single kernel implementation can handle different tensor shapes through parameterization.

**Rules:**

1. **One file per operation pattern**: If the same fused operation pattern (e.g., Linear+ReLU, Conv2d+ReLU+MaxPool) appears multiple times in the original model, create ONE subkernel file and instantiate it multiple times in `ModelNew.__init__` with different parameters.
2. **Parameterize by dimensions**: The subkernel `Model.__init__` should accept dimension arguments (e.g., `in_features`, `out_features`, `in_channels`, `out_channels`) that configure the operation for different shapes.
3. **Identify operation patterns, not instances**: When analyzing the forward pass, group by operation TYPE (e.g., "Linear+ReLU", "Conv2d+ReLU"), not by layer index or specific dimension values.
4. **get_inputs() shows ONE example shape**: The `get_inputs()` function should return tensors for one valid input configuration.
5. **get_input_configs() shows MULTIPLE example shapes**: If the same fused operation pattern appears multiple times in the original model, the file corresponding to the pattern must have  `get_input_configs()` function that returns a list of dictionaries with functions `get_inputs` and `get_init_inputs` for all shapes

**EXAMPLE - Applying the Reuse Principle:**

Consider a 4-layer MLP: `Linear(784,256) → GELU → Linear(256,128) → GELU → Linear(128,64) → GELU → Linear(64,10)`
This model has the pattern `[Linear+GELU] × 3` followed by `[Linear] × 1`.

**WRONG approach (creates duplicate files for the same operation):**
```
subkernel_problems/
    linear_gelu_1.py  # Linear(784, 256) + GELU
    linear_gelu_2.py  # Linear(256, 128) + GELU  ← DUPLICATE PATTERN!
    linear_gelu_3.py  # Linear(128, 64) + GELU   ← DUPLICATE PATTERN!
    linear_out.py     # Linear(64, 10)
```

**CORRECT approach (reuses the same class for identical operation patterns):**
```
subkernel_problems/
    linear_gelu.py    # Model(in_features, out_features) - ONE file, instantiated 3 times
    linear.py         # Model(in_features, out_features) - for final layer without activation
```

**Apply this principle to ANY model you decompose**: identify the unique operation patterns, create one subkernel file per pattern, and instantiate each pattern as many times as needed with appropriate dimension parameters.

## FUSION EXAMPLES

**Example 1: GELU + GEMM**
Input: `gelu(A) @ B`
→ Single submodule: GELU and matmul can be fused (gelu feeds directly into matmul)

**Example 2: Two-layer MLP**
Input: `relu(linear1(x)) → linear2`
→ Two submodules: 
  - Submodule 1: linear1 + relu (fused)
  - Submodule 2: linear2
  - Reason: Output of linear1+relu must be materialized before linear2

**Example 3: Attention Block**
Input: `softmax(Q @ K^T / sqrt(d)) @ V`
→ Depends on implementation:
  - If online softmax possible: single submodule
  - If standard softmax: split at softmax (requires full row before normalization)

## PROBLEM TO DECOMPOSE
```python
{{ problem_descriptor }}
```
\end{Verbatim}

\subsection{Subkernel Generation}

\begin{Verbatim}[breaklines=true]
**TASK: Generate a complete kernel implementation for the problem described in the "PYTORCH TO KERNEL PROBLEM" section.**


## PYTORCH TO KERNEL PROBLEM

You are given the following Triton kernel writing problem that contains
1. PyTorch module named `Model` that extends `torch.nn.Module`
2. Function that returns a sample argument for the module's `__init__` method `get_init_inputs`
3. Function that returns a sample argument for the module's `forward` method `get_inputs`


{{ problem_descriptor }}


Your task is to produce a PyTorch module named `ModelNew` that is a Triton implementation of the `Model` module satisfying the following requirements.
1. The `ModelNew` module contains `__init__` and `forward` methods with the same input and output signatures as the `Model` module
2. The `ModelNew` module's `forward` method must
  - Correctly and efficiently implement the `Model` module's `forward` method in Triton
  - Invoke a separately written Triton kernel function decorated with @triton.jit
    - The Triton kernel can be named anything (e.g., _kernel)
    - The Triton kernel function must explicitly accept all necessary tensors as function inputs rather than internally initializing random tensors
  - Handle grid calculation and kernel launch within the method, outside the Triton kernel function
  - Return contiguous tensors if returning tensors
3. If the `Model`'s `__init__` method implicitly initializes parameters/tensors, explicitly initialize these values in the exact same order in the `__init__` method
  - Example 1: `Model` contains `nn.Linear`/`nn.Conv3D` and its weights and bias values need to be explicitly passed to the Triton kernel function
    ```python
    # This is how nn.Linear initializes its weights and bias
    from torch.nn import init

    weight = <INITIALIZE APPROPRIATELY SHAPED EMPTY WEIGHT TENSOR ON THE CPU>
    init.kaiming_uniform_(weight, a=5**0.5)
    bias = <INITIALIZE APPROPRIATELY SHAPED EMPTY BIAS TENSOR ON THE CPU>
    fan_in, _ = init._calculate_fan_in_and_fan_out(weight)
    bound = 1 / fan_in**0.5 if fan_in > 0 else 0
    init.uniform_(bias, -bound, bound)
    ```
  - Example 2: `Model` contains `nn.MultiheadAttention` and its various parameters need to be explicitly passed to the Triton kernel function
    ```python
    from torch.nn import init

    <ASSUME APPROPRIATELY SHAPED TENSORS AND BOOLEAN FLAGS ARE INITIALIZED>
    if _qkv_same_embed_dim:
        init.xavier_uniform_(in_proj_weight)
    else:
        init.xavier_uniform_(q_proj_weight)
        init.xavier_uniform_(k_proj_weight)
        init.xavier_uniform_(v_proj_weight)

    if in_proj_bias is not None:
        init.constant_(in_proj_bias, 0.0)
        init.constant_(out_proj.bias, 0.0)
    if bias_k is not None:
        init.xavier_normal_(bias_k)
    if bias_v is not None:
        init.xavier_normal_(bias_v)
    ```
4. The `ModelNew` module must not cheat by initializing torch.nn module or trying to replicate `Model`'s forward pass using PyTorch operations or shortcuts
5. Do not repeat the raw original `Model` code anywhere, even in the comments
6. Keep logic in the `forward` method and the kernel minimal; speed is important and the entire `forward` method execution time is measured
7. Assume that the modules will be in PyTorch `eval` mode (this is highly relevant for BatchNorm, etc)
{% if assert_single_launch %}
8. The `ModelNew` module must only launch exactly ONE fused Triton kernel (multiple launches of the same kernel allowed) that performs all meaningful computation of the module
{% endif %}

{% if show_example -%}
Example solution format for Triton persistent matmul kernel (wrapper can use PyTorch, but kernel must purely use Triton):
```python
import torch
import triton
import triton.language as tl

@triton.jit
def _compute_pid(tile_id, num_pid_in_group, num_pid_m, GROUP_SIZE_M, NUM_SMS):
    group_id = tile_id // num_pid_in_group
    first_pid_m = group_id * GROUP_SIZE_M
    group_size_m = min(num_pid_m - first_pid_m, GROUP_SIZE_M)
    pid_m = first_pid_m + (tile_id % group_size_m)
    pid_n = (tile_id % num_pid_in_group) // group_size_m
    return pid_m, pid_n


@triton.autotune(
    configs=matmul_get_configs(),
    key=["M", "N", "K"],
)
@triton.jit(launch_metadata=_matmul_launch_metadata)
def matmul_kernel_persistent(a_ptr, b_ptr, c_ptr,
                             M, N, K,
                             stride_am, stride_ak,
                             stride_bk, stride_bn,
                             stride_cm, stride_cn,
                             BLOCK_SIZE_M: tl.constexpr,
                             BLOCK_SIZE_N: tl.constexpr,
                             BLOCK_SIZE_K: tl.constexpr,
                             GROUP_SIZE_M: tl.constexpr,
                             NUM_SMS: tl.constexpr,
                             ):
    start_pid = tl.program_id(axis=0)
    num_pid_m = tl.cdiv(M, BLOCK_SIZE_M)
    num_pid_n = tl.cdiv(N, BLOCK_SIZE_N)
    k_tiles = tl.cdiv(K, BLOCK_SIZE_K)
    num_tiles = num_pid_m * num_pid_n

    # NOTE: There is currently a bug in blackwell pipelining that means it can't handle a value being
    # used in both the prologue and epilogue, so we duplicate the counters as a work-around.
    tile_id_c = start_pid - NUM_SMS

    offs_k_for_mask = tl.arange(0, BLOCK_SIZE_K)
    num_pid_in_group = GROUP_SIZE_M * num_pid_n

    for tile_id in tl.range(start_pid, num_tiles, NUM_SMS, flatten=True):
        pid_m, pid_n = _compute_pid(tile_id, num_pid_in_group, num_pid_m, GROUP_SIZE_M, NUM_SMS)
        start_m = pid_m * BLOCK_SIZE_M
        start_n = pid_n * BLOCK_SIZE_N
        offs_am = start_m + tl.arange(0, BLOCK_SIZE_M)
        offs_bn = start_n + tl.arange(0, BLOCK_SIZE_N)
        offs_am = tl.where(offs_am < M, offs_am, 0)
        offs_bn = tl.where(offs_bn < N, offs_bn, 0)
        offs_am = tl.max_contiguous(tl.multiple_of(offs_am, BLOCK_SIZE_M), BLOCK_SIZE_M)
        offs_bn = tl.max_contiguous(tl.multiple_of(offs_bn, BLOCK_SIZE_N), BLOCK_SIZE_N)

        accumulator = tl.zeros((BLOCK_SIZE_M, BLOCK_SIZE_N), dtype=tl.float32)
        for ki in range(k_tiles):
            offs_k = ki * BLOCK_SIZE_K + tl.arange(0, BLOCK_SIZE_K)
            a_ptrs = a_ptr + (offs_am[:, None] * stride_am + offs_k[None, :] * stride_ak)
            b_ptrs = b_ptr + (offs_k[:, None] * stride_bk + offs_bn[None, :] * stride_bn)

            a = tl.load(a_ptrs, mask=offs_k_for_mask[None, :] < K - ki * BLOCK_SIZE_K, other=0.0)
            b = tl.load(b_ptrs, mask=offs_k_for_mask[:, None] < K - ki * BLOCK_SIZE_K, other=0.0)
            accumulator = tl.dot(a, b, accumulator)

        tile_id_c += NUM_SMS
        pid_m, pid_n = _compute_pid(tile_id_c, num_pid_in_group, num_pid_m, GROUP_SIZE_M, NUM_SMS)
        offs_cm = pid_m * BLOCK_SIZE_M + tl.arange(0, BLOCK_SIZE_M)
        offs_cn = pid_n * BLOCK_SIZE_N + tl.arange(0, BLOCK_SIZE_N)
        c_ptrs = c_ptr + stride_cm * offs_cm[:, None] + stride_cn * offs_cn[None, :]
        c_mask = (offs_cm[:, None] < M) & (offs_cn[None, :] < N)
        if (c_ptr.dtype.element_ty == tl.float8e4nv):
            c = accumulator.to(tl.float8e4nv)
        else:
            c = accumulator.to(tl.float32)
        tl.store(c_ptrs, c, mask=c_mask)


class ModelNew(torch.nn.Module):
    def __init__(self, *args, **kwargs):
        """Any necessary module initialization logic."""
        ...

    def forward(self, a, b):
        """Wrapper function that handles kernel launch."""
        assert a.shape[1] == b.shape[0], "Incompatible dimensions"
        assert a.dtype == b.dtype, "Incompatible dtypes"
        NUM_SMS = torch.cuda.get_device_properties("cuda").multi_processor_count
        M, K = a.shape
        K, N = b.shape
        dtype = a.dtype
        # Allocates output. NOTE: PyTorch operations allowed here only for setup
        c = torch.empty((M, N), device=a.device, dtype=dtype)
        # 1D launch kernel where each block gets its own program.
        grid = lambda META: (min(NUM_SMS, triton.cdiv(M, META["BLOCK_SIZE_M"]) * triton.cdiv(N, META["BLOCK_SIZE_N"])), )
        matmul_kernel_persistent[grid](
            a, b, c,
            M, N, K,
            a.stride(0), a.stride(1),
            b.stride(0), b.stride(1),
            c.stride(0), c.stride(1),
            NUM_SMS=NUM_SMS,
        )
        return c
```
{%- endif %}

Your solution will be assessed in two ways.
1. Correctness: Are the outputs of `Model` and `ModelNew`'s `forward` methods sufficiently close?
2. Speedup: How fast is the `ModelNew`'s Triton-based `forward` method compared to `Model`'s PyTorch-based `forward` method?

## General Triton Guidelines (Helpful):

1. KERNEL STRUCTURE:
  - Use @triton.jit decorator for kernel functions
  - Use tl.constexpr for compile-time constants (BLOCK_SIZE, etc.)
  - Include proper type hints and launch metadata when needed

2. MEMORY ACCESS PATTERNS:
  - Use tl.load and tl.store with proper masking
  - Coalesce memory accesses for optimal performance
  - Use tensor descriptors for advanced memory operations (TMA)
  - Handle boundary conditions with masks

3. INDEXING AND GRID:
  - Use tl.program_id(axis) for block indices
  - Calculate offsets: pid * BLOCK_SIZE + tl.arange(0, BLOCK_SIZE)
  - Use tl.cdiv for ceiling division
  - Always mask for out-of-bounds protection

4. OPTIMIZATION TECHNIQUES:
  - Use @triton.autotune for automatic configuration selection (Autotuned parameters must be declared as tl.constexpr meta-parameters and must not be passed as runtime kernel arguments)
  - Choose appropriate BLOCK_SIZE (powers of 2: 64, 128, 256, 512, 1024)
  - Leverage tensor cores with tl.dot for matrix operations
  - Use warp specialization for better scheduling
  - Consider epilogue subtiling to reduce shared memory usage
  - Aggressively fuse compatible operator stages to minimize memory traffic and kernel launch overhead; only keep stages separate if fusion is infeasible

5. COMMON PATTERNS:
  a) Elementwise operations: Load -> Compute -> Store
  b) Reductions: Use tl.reduce with proper axis (can also use tl.sum(), tl.max(), tl.min(), etc)
  c) Matrix multiplication: Use tl.dot with accumulator
  d) Softmax: Online normalization for numerical stability
  e) Fused operations: Combine multiple ops in single kernel and document the fused stages for reviewers
    - BN inside fused conv:
      - bn_mean_f32 = tl.load(mean_ptr + oc).to(tl.float32)
      - bn_var_f32 = tl.load(var_ptr + oc).to(tl.float32)
      - gamma_f32 = tl.load(gamma_ptr + oc).to(tl.float32)
      - beta_f32  = tl.load(beta_ptr + oc).to(tl.float32)
      - acc_f32   = acc.to(tl.float32)  # conv accumulator
      - x_norm    = (acc_f32 - bn_mean_f32) / tl.sqrt(bn_var_f32 + eps)
      - out_f32   = x_norm * gamma_f32 + beta_f32
      - tl.store(out_ptr + out_idx, out_f32.to(out_ptr.dtype.element_ty), mask=mask)
    - LayerNorm per row:
      - mean_f32 = tl.zeros((), dtype=tl.float32)
      - var_f32  = tl.zeros((), dtype=tl.float32)
      - accumulate in fp32; use tl.sqrt(var_f32 + eps); cast on store.

6. ADVANCED FEATURES:
  - Persistent kernels for better SM utilization
  - Tensor Memory Accelerator (TMA) descriptors
  - Multi-stage pipelines with num_stages
  - Warp specialization with warp_specialize parameter

7. RUNTIME CONSTRAINTS:
  - Wrappers: validate/allocate/launch only; no math
  - All compute runs in Triton kernels; no torch.nn, torch.nn.functional (e.g., F.*), or other PyTorch compute ops, including general tensor-tensor math like torch.matmul/mm/bmm/einsum or their Tensor method forms, and no low-level torch.ops.aten.* calls (e.g., torch.ops.aten.conv2d, torch.ops.aten.layer_norm, torch.ops.aten.addmm, torch.ops.aten.mean)
  - PyTorch allowed only for allocation, dtype/device checks, and packaging results

8. TRITON MATH DOCS (from `triton.language` module or `tl`):
  - The following is a comprehensive list of math related `tl` ops; if a math op in `tl` is not in the following list, it does not exist
    - LinAlg Ops: tl.dot, tl.dot_scaled
    - Math Ops: tl.abs, tl.cdiv, tl.ceil, tl.clamp, tl.cos, tl.div_rn, tl.erf, tl.exp, tl.exp2, tl.fdiv, tl.floor, tl.fma, tl.log, tl.log2, tl.maximum, tl.minimum, tl.rsqrt, tl.sigmoid, tl.sin, tl.softmax, tl.sqrt, tl.sqrt_rn, tl.umulhi
    - Reduction Ops: tl.argmax, tl.argmin, tl.max, tl.min, tl.reduce, tl.sum, tl.xor_sum

## Basic Hardware Information

The Triton kernel should be optimized for NVIDIA A100 GPU with 80GB RAM and Ampere architecture.

GPU Details
- SMs: 108 (SXM) / 80 (PCIe)
- Warp size: 32 threads
- Max threads / block 1024 → `num_warps <= 32`
- Registers / SM: 256k 32-bit → large tiles easily become register-bound
- Shared memory / SM: up to 164 KB (configurable)
- L2 cache* ~40 MB (huge → favors block reuse / grouping)
- Tensor Cores**: FP16, BF16, TF32 (TF32 enabled by default in cuBLAS)

Important Suggestions
- Always set `tl.dot`'s `allow_tf32` flag to False; otherwise correctness tests will fail

**RUNTIME RESTRICTIONS:**
- The Python wrapper may only perform argument validation, tensor allocation, and launch configuration
- All math (convs, activations, pooling, reductions, etc.) must reside inside kernels
- Never import or instantiate `torch.nn` modules within the main kernel function, call `torch.nn.functional` (including aliases like `F.*`), or rely on PyTorch helpers such as `torch.conv*`, `torch.relu`, `torch.max_pool*`, etc
- PyTorch usage is limited to allocation helpers, dtype/device checks, and assertions needed to launch the main kernel function

**STRICTLY FORBIDDEN - DO NOT CHEAT:**
- DO NOT call PyTorch functions (torch.add, torch.mul, torch.sum, torch.matmul, torch.mm, torch.bmm, torch.einsum, etc.) inside the kernel computation
- DO NOT use PyTorch operations to perform the actual computation and just return the result (this includes tensor-tensor ops such as x.matmul(y), x.bmm(y), x.mm(y), or x.einsum(...))
- DO NOT call low-level operator APIs like torch.ops.aten.* (e.g., torch.ops.aten.conv2d, torch.ops.aten.addmm, torch.ops.aten.layer_norm, torch.ops.aten.mean); these are PyTorch compute ops and using them is considered cheating
- DO NOT implement the logic using pure PyTorch and avoid writing kernel code
- DO NOT hide PyTorch compute (including torch.ops.aten.* calls or torch.nn/torch.nn.functional usage) inside helper functions that are then called from the wrapper
- The wrapper class/method can use PyTorch for tensor creation, memory allocation, and result formatting, but the core computation MUST happen in the main kernel function
- DO NOT import or instantiate torch.nn modules, call torch.nn.functional (including aliases like F.*), or use PyTorch activations/pooling helpers to satisfy the requirements


## FINAL CHECKLIST

**Runtime Constraints (CRITICAL: Evaluation will FAIL if these are not satisfied)**

* [ ] Wrapper PyTorch module `ModelNew` that validates, allocates, and launches kernel function(s) (no math) exists in the code
* [ ] No other PyTorch module class definitions exists in the code, including the problem PyTorch module `Model`, exists in the code
* [ ] Kernel function(s) use @triton.jit decorator
* [ ] All compute runs in Triton kernels
* [ ] No torch.nn, torch.nn.functional (F.*),torch.ops.aten.*, torch.matmul/mm/bmm/einsum (and their Tensor method equivalents), or PyTorch compute ops
* [ ] PyTorch used only for allocation, dtype/device checks, and packaging results
* [ ] Explicitly reason about what is being fused and not fused inside the wrapper docstring

**Kernel Structure**

* [ ] Compile-time constants use tl.constexpr (BLOCK_SIZE, etc.)
* [ ] Proper type hints and launch metadata included
* [ ] Kernel is self-contained and stateless

**Memory Access**

* [ ] tl.load and tl.store used with proper masking
* [ ] Memory accesses are coalesced (inner dimension maps to adjacent threads)
* [ ] Boundary conditions handled with masks
* [ ] Invariants loaded once and reused from registers

**Indexing & Grid**

* [ ] tl.program_id(axis) used for block indices
* [ ] Offsets calculated as: pid * BLOCK_SIZE + tl.arange(0, BLOCK_SIZE)
* [ ] Out-of-bounds protection with masks

**Parallelism & Launch**

* [ ] BLOCK_SIZE is power of 2 (64, 128, 256, 512, 1024)
* [ ] @triton.autotune used for automatic configuration selection

Fill out the following solution template and present it as a single Python code block:

```python
import triton
import triton.language as tl
import torch.nn as nn

# TODO: Write kernel function(s) wrapped in @triton.jit here


# TODO: Invoke the kernel function(s) from above *inside a PyTorch module named ModelNew* to solve the problem
class ModelNew(nn.Module):
    ...
```
\end{Verbatim}


\end{document}